\documentclass{article}

\usepackage{arxiv}

\usepackage{graphicx}

\usepackage{tikz}
\usepackage{comment} 
\usepackage{amsmath,amssymb} 
\usepackage{color}

\usepackage{placeins}
\usepackage{xparse}
\usepackage{etoolbox}

\usepackage{algorithm}
\usepackage{algpseudocode}
\usepackage{hyperref}

\newcommand{\fig}[1]{Figure~\ref{fig:#1}}

\newcommand{\tab}[1]{Table~\ref{tab:#1}}

\begin{document}
\pagestyle{headings}
\def\ECCVSubNumber{4302}  

\title{DeepLandscape: Adversarial Modeling of Landscape Videos} 


\author{ {Elizaveta Logacheva} \\
	Samsung AI Center, Moscow\\
	\texttt{elimohl@gmail.com} \\
	\And
	{Roman Suvorov} \\
	Samsung AI Center, Moscow\\
	\And
	{Oleg Khomenko} \\
	Samsung AI Center, Moscow\\
	\And
	{Anton Mashikhin} \\
	Samsung AI Center, Moscow\\
	\And
	{Victor Lempitsky} \\
	Samsung AI Center, Moscow\\
	Skolkovo Institute of Science and Technology, Moscow
}

%

\maketitle

\begin{abstract}
We build a new model of landscape videos that can be trained on a mixture of static landscape images as well as landscape animations. Our architecture extends StyleGAN model by augmenting it with parts that allow to model dynamic changes in a scene. Once trained, our model can be used to generate realistic time-lapse landscape videos with moving objects and time-of-the-day changes. Furthermore, by fitting the learned models to a static landscape image, the latter can be reenacted in a realistic way. We propose simple but necessary modifications to StyleGAN inversion procedure, which lead to in-domain latent codes and allow to manipulate real images. Quantitative comparisons and user studies suggest that our model produces more compelling animations of given photographs than previously proposed methods. The results of our approach including comparisons with prior art can be seen in supplementary materials and on the project page \url{https://saic-mdal.github.io/deep-landscape/}.
\end{abstract}

\section{Introduction}
\label{sect:intro}

\begin{figure}
    \centering
    \includegraphics[width=\linewidth]{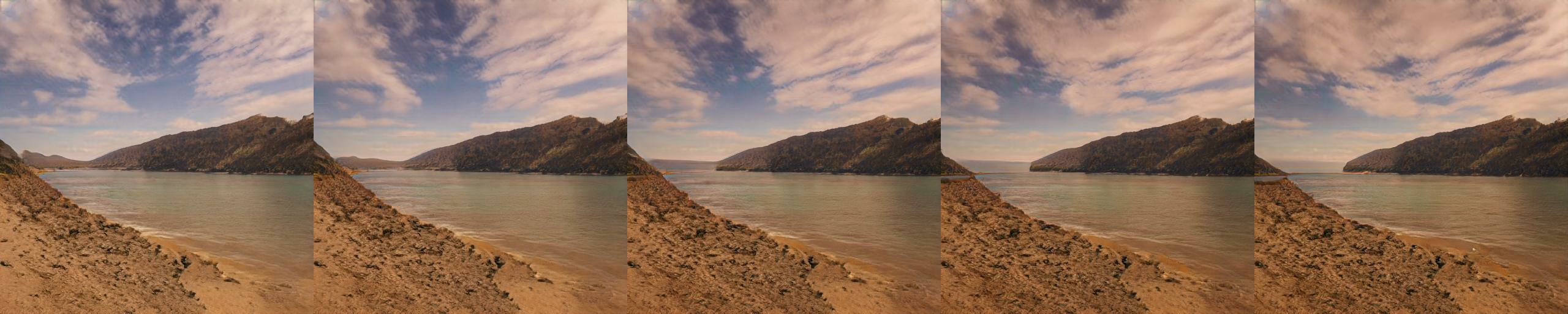}
    
    \includegraphics[width=\linewidth]{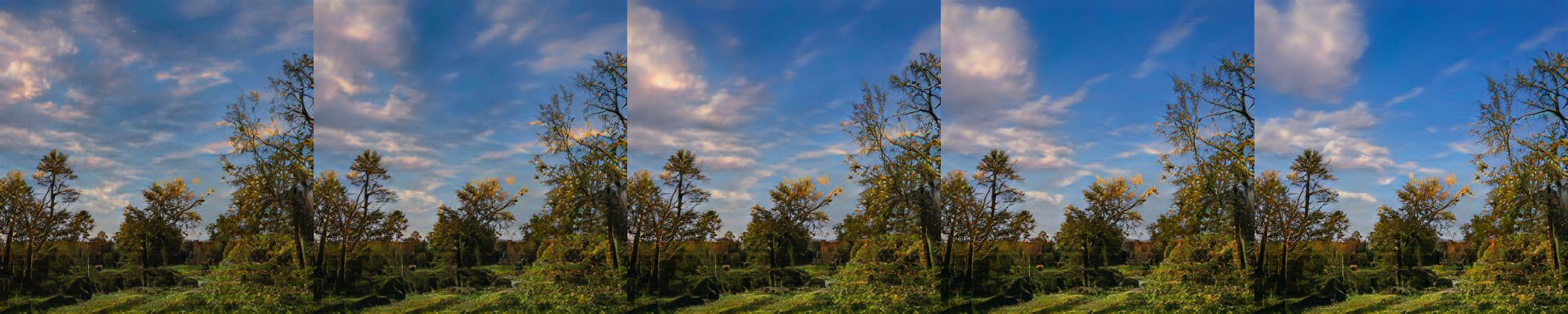}

    \includegraphics[width=\linewidth]{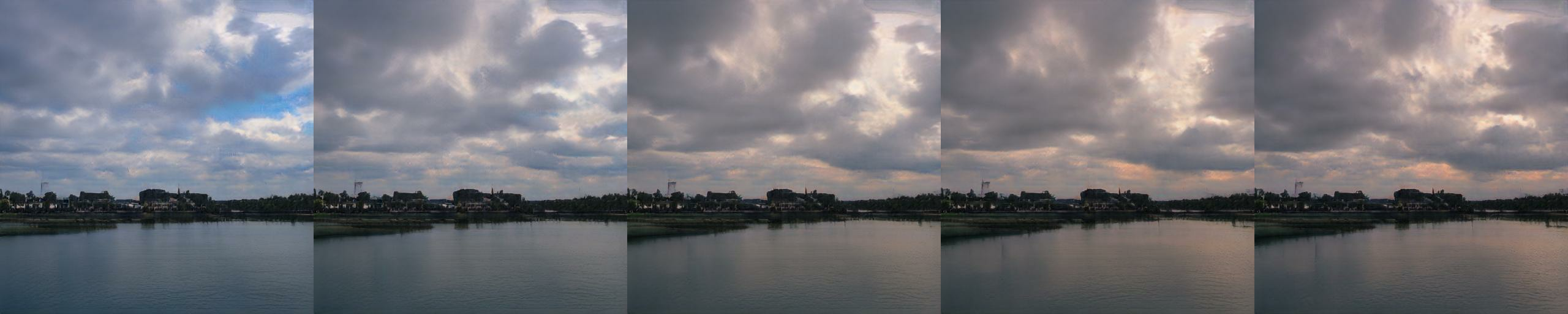}
    
    \includegraphics[width=\linewidth]{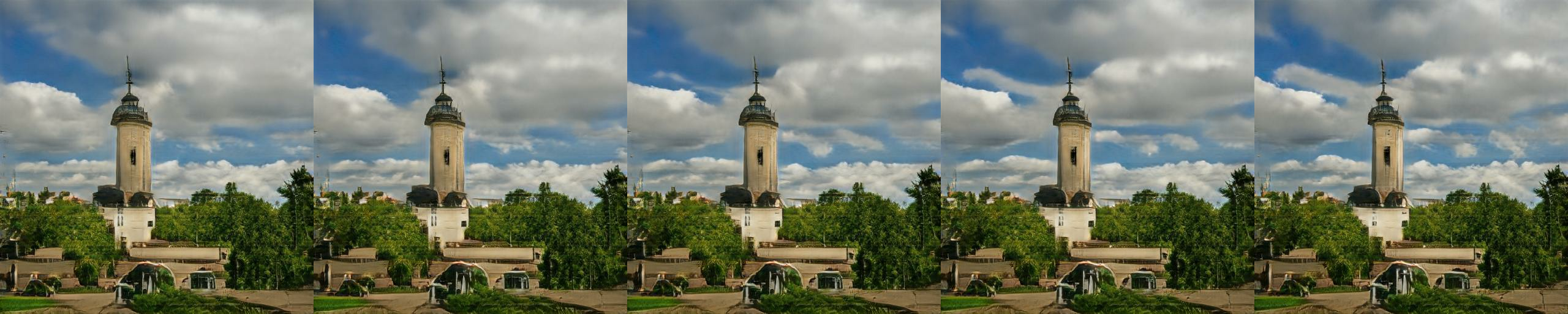}
    
    \caption{Videos generated by the DeepLandscape model. Each row shows a separate video, obtained by sampling the static and dynamic components randomly, and then animating the dynamic components using homography warping. These videos are generated at $512 \times 512$ resolution (\textit{zoom-in recommended}).}
    \label{fig:synth}
\end{figure}

This work is motivated by the ``bringing landscape images to life'' application. We thus aim to build a system that for a given landscape photograph, generates its plausible animation with realistic movements and global lighting changes. To achieve our goal, we first build a generative model (\fig{synth}) of timelapse landscape videos, which can successfully capture complex aspects of this domain. These complexities include both static aspects such as abundance of spatial details, high variability of texture and geometry, as well as dynamic complexity including motions of clouds, waves, foliage, and global lighting changes. 
We build our approach upon the recent progress in the generative modeling of images, and specifically the StyleGAN model~\cite{Karras2018ASG}.
We show how to change the StyleGAN model to learn and to decompose different dynamic effects: global changes are controlled by the non-convolutional variables, strong local motions are controlled by “noise branch” inputs.

Similarly to the original StyleGAN model, ours requires a large amount of training data. While it is very hard to obtain a very large dataset of high-quality scenery timelapse videos, obtaining a large-scale dataset of scenery static images is much easier. We thus suggest how our generative model can be learned from two sources, namely (i) a large-scale dataset of static images, (ii) a smaller dataset of videos. Previous video GANs learn motion from sequences of consecutive video frames. We show that learning on randomly taken frames without an explicit motion model is possible. It allows to disentangle static appearance from the dynamic, as well as manifold of possible changes from a trajectory in it. 

Once trained, our model can animate a given photograph. We first fit the latent variables of the model to the provided image, and then obtain the animation by changing the subset of variables corresponding to dynamic aspects appropriately. As our model has more latent parameters than a given static image, fitting them to a photograph is an ill-posed problem, and we develop a particular method for such fitting that results in plausible animations. While our model is trained to generate images at medium resolution (256$\times$256 or 512$\times$512 
), we show that we can postprocess the results with an appropriately trained super-resolution network to obtain videos at higher resolution (up to one megapixel).

In the experiments, we assess the realism of synthetic videos sampled from our generative model and its ablations. Furthermore, we evaluate our approach at our main task (``bringing landscape images to life''). For this task, both quantitative comparisons and, more importantly, user studies reveal a significant advantage of our system over the three recently proposed approaches \cite{Endo2019AnimatingLS,Shaham2019SinGANLA,Tesfaldet2017TwoStreamCN}.


\section{Related work}
\label{sect:related}

Learning video representation and predicting future frames using deep neural networks is a very active area of research~\cite{Srivastava2015UnsupervisedLO,Villegas2017DecomposingMA,Mathieu2015DeepMV,finn2016unsupervised}. Most early works are focused on using deep neural networks (DNNs) with recurrent units (GRU or LSTM) and train them in supervised manner to obtain next frame using pixel-level prediction \cite{finn2016unsupervised,Srivastava2015UnsupervisedLO}. At the same time, Generative Adversarial Nets (GANs)~\cite{Goodfellow2014GenerativeAN} have achieved very impressive results for image generation, and recently several methods extending them to video have been suggested. Some GAN-based models consider single image as an input (\textit{image2video})~\cite{pan2019video,Li2018FlowGroundedSV}, while others input sequences of frames (\textit{video2video}, ~\cite{Mathieu2015DeepMV,wang2018vid2vid,aigner2018futuregan,Xiong_2018_CVPR,Li2019FromHT}). In this work we focus only on the image2video setting. Training GANs for video-generation often performed with two discriminator networks: single image and temporal discriminators~\cite{wang2018vid2vid,Clark2019EfficientVG,Xiong_2018_CVPR}. In this work we propose to use a simplified temporal discriminator, which only looks at unordered pairs of frames.

Video generation/prediction works generally consider either videos with articulated objects/multiple moving objects~\cite{soomro2012ucf101,Carreira2018ASN} or videos with weakly structured moving objects or dynamic textures such as clouds, grass, fire~\cite{Xiong_2018_CVPR,Tesfaldet2017TwoStreamCN}.
Our work is more related to the latter case, namely: landscape photos and videos. Because of the domain specifics, we can model spatial motions in the video in the \textit{latent} space using simple homography transformations, and let the generator to synthesize plausible deviations from this simplistic model. Our approach is thus opposed to methods that animate landscapes and textures by generating warping fields applied to the \textit{raw pixels} of the input static image~\cite{Endo2019AnimatingLS,Chen2017VideoIF,Li2018FlowGroundedSV,van2017transformation,Chuang2005}. 
Animation in the latent space as well as the separation of latent space into static and dynamic components has been proposed and investigated in~\cite{Tulyakov2017MoCoGANDM,Villegas2017DecomposingMA,Vondrick2016GeneratingVW,Endo2019AnimatingLS,denton2017unsupervised}. Our work modifies and extends these ideas to the StyleGAN~\cite{Karras2018ASG} model.

As we need to find latent space embedding of static images in order to animate them, we follow a number of works on GAN inversion (inference). Here, we borrow ideas of using an encoder into the latent space followed by gradient descent~\cite{Zhu16}, the latent space expansion for StyleGAN~\cite{Abdal2019Image2StyleGANHT}, and generator fine-tuning~\cite{Bau19,zakharov2019few}. On top of that, we have to make several important adjustments to the inference procedure specific to our architecture, and we show that without such adjustments the animation works poorly.

\section{Method}
\label{sect:method}

%

\newcommand{\wow}[1][]{\ifblank{#1}{}{,#1}}

\newcommand{\zs}{\mathbf{z}^\text{st}}  
\newcommand{\zd}[1][]{\mathbf{z}^\text{dyn\wow[#1]}}  
\newcommand{\zall}{\mathbf{z}} 
\newcommand{\ds}{D^\text{st}}  
\newcommand{\dd}{D^\text{dyn}}  
\newcommand{\sns}[1]{S^\text{st}_{#1}}  
\newcommand{\snd}[2][]{S^\text{dyn}_{#2\wow[#1]}}  
\newcommand{\sn}[1]{S_{#1}}  
\newcommand{\snsh}[1]{\hat{S}^\text{st}_{#1}}  
\newcommand{\sndh}[1]{\hat{S}^\text{dyn}_{#1}}  
\newcommand{\snh}[1]{\hat{S}_{#1}}  
\newcommand{\Sall}{{\cal S}}
\newcommand{\Sd}[1][]{{\cal S}^\text{dyn\wow[#1]}}
\newcommand{\Ss}{{\cal S}^\text{st}}

\newcommand{\w}{\mathbf{w}}
\newcommand{\Wall}{{\cal W}}
\newcommand{\nl}{N}
\newcommand{\maxres}{2^{\nl + 1}}

\newcommand{\M}{\mathbf{M}} 
\newcommand{\G}{\mathbf{G}} 
\newcommand{\E}{\mathbf{E}} 
\newcommand{\A}{\mathbf{A}} 

\newcommand{\realvec}[1]{\mathbb{R}^{#1}}
\newcommand{\realsq}[1]{\mathbb{R}^{#1 \times #1}}
\newcommand{\realmat}[2]{\mathbb{R}^{#1 \times #2}}

\newcommand{\V}{{\cal V}} 
\newcommand{\I}{{\cal I}} 

\subsection{Generative model of timelapse videos}

\subsubsection{Model architecture.}

\begin{figure}
    \centering
    \includegraphics[angle=270,width=\textwidth]{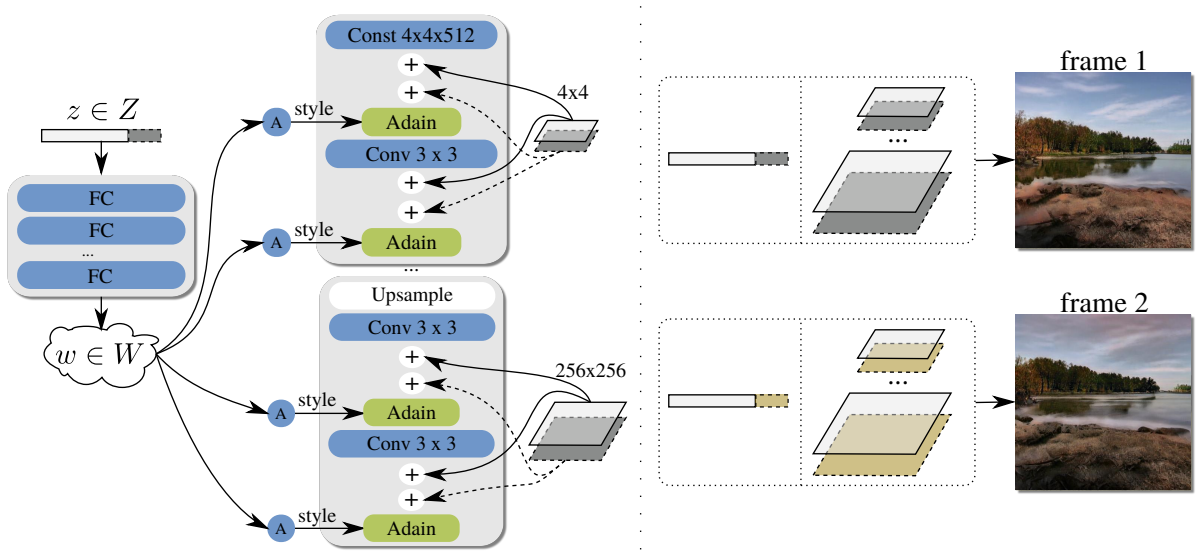}
    \caption{Left -- the generator used by our model (augmented StyleGAN generator). The main difference from StyleGAN is the second set of spatial input tensors (darkgray). Right -- sampling procedure for our model. Two frames of the same video can be sampled by using same static latent variables (lightgray), and two different sets of dynamic latent variables (darkgray and yellow). }
    \label{fig:diagram}
\end{figure} 

The architecture of our model is based on StyleGAN~\cite{Karras2018ASG}. Our model outputs images of resolution $256\times{}256$ (or $512\times{}512$) and has four sets of latent variables:
\begin{itemize}
    \item a vector $\zs \in \realvec{\ds}$, which encodes colors and the general scene layout;
    \item a vector $\zd \in \realvec{\dd}$, which encodes global lighting (e.g.\ time of day);
    \item a set $\Ss$ of square matrices $\sns{1} \in \realsq{4}$, ..., $\sns{\nl} \in \realsq{\maxres}$, which encode shapes and details of static objects at $\nl=7$ different resolutions between $4\times 4$ and $256\times 256$ ($\nl=8$ for $512\times 512$);
    \item a set $\Sd$ of square matrices $\snd{1} \in \realsq{4}$, ..., $\snd{\nl} \in \realsq{\maxres}$, which encode shapes and details of dynamic objects at the corresponding resolutions.
\end{itemize}

Our generator has two components: the multilayer perceptron $\M$ and the convolutional generator $\G$. As in \cite{Karras2018ASG}, the perceptron $\M$ takes the concatenated vector  $\zall = \left[ \zs,\, \zd \right] \in \realvec{512}$ and transforms it to the \textit{style vector} $\w \in \realvec{512}$. The convolutional generator $\G$ also follows~\cite{Karras2018ASG} and has $\nl=7$ (or 8)  blocks. Within each block, a convolution is followed by two elementwise additions of two tensors obtained from $\sns{n}$ and $\snd{n}$ by a learnable per-channel scaling (whereas \cite{Karras2018ASG} has only one addition). Finally, the AdaIN~\cite{huang2017arbitrary} transform is applied using per-channel scales and biases obtained from $\w$ using learnable linear transform. Within each block, this sequence of steps is repeated twice followed by upsampling and convolution layers. 

Below, we will refer to the set of input latent variables $$\left\{ \zs, \zd, \sns{1}, ..., \sns{\nl}, \snd{1}, ..., \snd{\nl} \right\}$$ as \textit{original inputs} (or original latents). As in StyleGAN, the convolutional generator may use \textit{separate} $\w$ vectors at each of the resolution (style mixing). We will then refer to the set of all style vectors as $\Wall = \left\{ \w_1, ..., \w_\nl \right\}$. Finally, we will denote the set of all spatial random inputs of the generator as $ \Sall = \{\Ss,\, \Sd\} = \left\{ \sns{1}, ..., \sns{\nl}, \snd{1}, ..., \snd{\nl} \right\}$.


\subsubsection{Learning the model.}
The model is trained from two sources of data, the dataset of static scenery images $\I$ and the dataset of timelapse scenery videos $\V$. It is relatively easy to collect a large static dataset, while with our best efforts we were able to collect a few hundreds of videos, that do not cover all the diversity of landscapes. Thus, both sources of data have to be utilized in order to build a good model. To do that, we train our generative model in an adversarial way with two different discriminators. 

The \textit{static discriminator} $D_{st}$ has the same architecture and design choises as in StyleGAN. It observes images from $\I$ as real, while the fake samples are generated by our model. The \textit{pairwise discriminator} $D_{dyn}$ looks at pairs of images. It duplicates the architecture of $D_{st}$ except first convolutional block that is applied separately to each frame. A real pair of images is obtained by sampling a video from $\V$, and then sampling two random frames (arbitrary far for each other) from it. A fake pair is obtained by sampling common static latents $\zs$ and $\Ss$, and then individual dynamic latents $\zd[1]$, $\zd[2]$ and $\Sd[1]$, $\Sd[2]$. The two images are then obtained as $\G(\M(\zs,\zd[1]),\,\Ss,\Sd[1])$ and $\G(\M(\zs,\zd[1]),\,\Ss,\Sd[2])$. All samples are drawn from unit normal distributions.


The model is trained within standard GAN approach with non-saturating loss~\cite{Goodfellow2014GenerativeAN} with R1 regularization~\cite{mescheder2018which} as in the original StyleGAN paper. During each update of the generator, we either sample a batch of fake images to which the static discriminator is applied or a batch of image pairs to which the pairwise discriminator is applied. The proportions of the static discriminator and the pairwise discriminator are annealed from 0.5/0.5 to 0.9/0.1 respectively over each resolution transition phase and then kept fixed at 0.1. This helps the generator to learn disentangle static and dynamic latents early for each resolution and prevents the pairwise generator from overfitting to our relatively small video dataset.


During learning, we want the pairwise discriminator to focus on the inconsistencies within each pair, and leave visual quality to the static discriminator. Furthermore, since the pairwise discriminator only sees real frames sampled from a limited number of videos, it may prone overfit to this limited set and effectively stop contributing to the learning process (while the static discriminator, which observes more diverse set of scenes, keeps improving the diversity of the model). It turns out, both problems (focus on image quality rather than pairwise consistency, overfitting to limited diversity of videos) can be solved with a simple trick.  We augment the fake set of frames with pairs of crops taken from same video frame, but from different locations. Since these crops have the same visual quality as the images in real frames, and since they come from the same videos as images within real pairs, the pairwise discriminator effectively stops paying attention to image quality, cannot simply overfit to the statistics of scenes in the video dataset, and has to focus on finding pairwise inconsistencies within fake pairs. We observed this \textit{crop sampling} trick to improve the quality of our model significantly.


\begin{figure}
    \centering
    \begin{tabular}{c|c c c c c c c}
        \hline\hline
        Config           & I2S~\cite{Abdal2019Image2StyleGANHT}
                                  & MO   & E      & EO   & EOI  & EOIF  & EOIFS \\
        \hline
        Init $\Wall$     & Mean   & Mean & $\E$   & $\E$ & $\E$ & $\E$  & $\E$   \\
        Init $\Sall$     & Random & Zero & Random & Zero & Zero & Zero  & Zero   \\
        Optimize $\Sall$ &        & +    &        & +    & +    & +     & +      \\
        Optimize $\Wall$ & +      & +    &        & +    & +    & +     & +      \\
        $L^O_{init}$     &        &      &        &      & +    & +     & +      \\
        Fine-Tune $\G$   &        &      &        &      &      & +     & +      \\
        Segmentation     &        &      &        &      &      &       & +      \\
        \hline
        \includegraphics[height=4.1em]{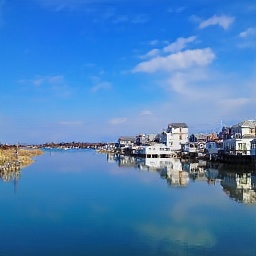} &
        \includegraphics[height=4.1em]{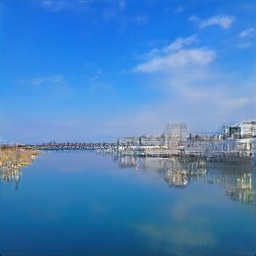}  &
        \includegraphics[height=4.1em]{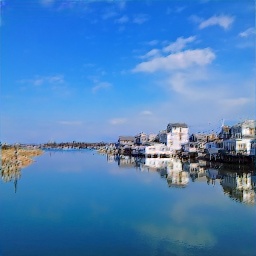} & 
        \includegraphics[height=4.1em]{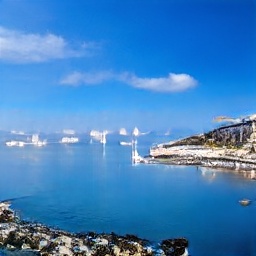} &
        \includegraphics[height=4.1em]{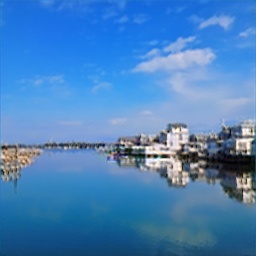} &
        \includegraphics[height=4.1em]{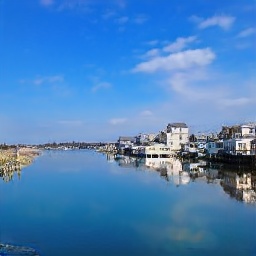} &
        \includegraphics[height=4.1em]{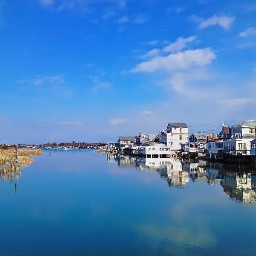} &
        \includegraphics[height=4.1em]{images/fig3/fcz3_ss_470_01_eoifs/00001.jpg}
        \\
        
         &
        \includegraphics[height=4.1em]{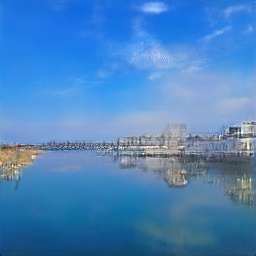}  &
        \includegraphics[height=4.1em]{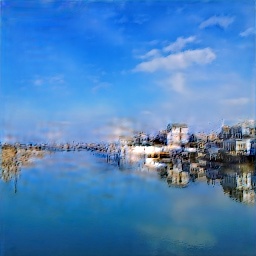} & 
        \includegraphics[height=4.1em]{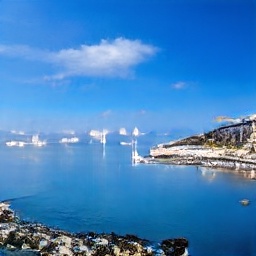} &
        \includegraphics[height=4.1em]{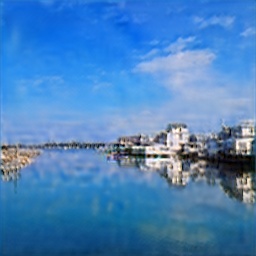} &
        \includegraphics[height=4.1em]{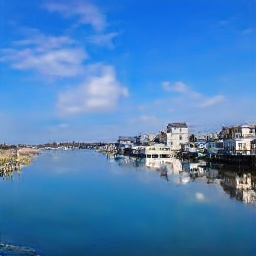} &
        \includegraphics[height=4.1em]{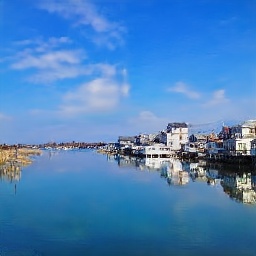} &
        \includegraphics[height=4.1em]{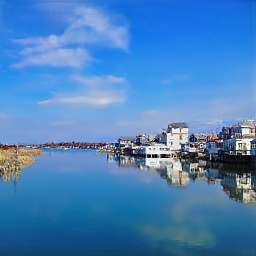}
        \\
        \hline
        Reconstruction & - & + & - & + & $\pm$ & + & + \\
        Animation      & - & - & + & - &     + & + & + \\
        \hline\hline
    \end{tabular}
    
    \caption{The effect of different inference algorithms on the reconstruction quality and the ability to animate. \textbf{Left column}: original image. \textbf{First row}: reconstructions obtained with different inference algorithms. \textbf{Second row}: a frame from animation ($\Sd$ are shifted 50\% left). Note that I2S~\cite{Abdal2019Image2StyleGANHT} does not work well in our case, since our generator relies on $\Sall$ more than the original StyleGAN method. $L^O_{init}$ is a regularization term applied to $\Wall$ during inference, which makes latents to stay in-domain and allows to manipulate real images. We quantify these effects in \ref{infquant}.}
    \label{fig:inference-comparison}
\end{figure}

\subsubsection{Sampling videos from the model.} 

Our model does not attempt to learn full temporal dynamics of videos, and instead focuses on pairwise consistency of frames that are generated when the dynamic latent variables are resampled. In particular, the pairwise discriminator in our model does not sample real frames sequentially. The sampling procedure for fake pairs does not try to generate adjacent frames either. One of the reasons why we do not attempt to learn continuity, is because the training dataset contains videos of widely-varying temporal rates, making the notion of temporal adjacency for a pair of frames effectively meaningless.

Because of this our generation process is agnostic to a model of motion. The generator is forced to produce plausible frames regardless of $\Sd$ and $\zd$ changes.
In our experiments we found that a simple model of motion described below is enough to produce compelling videos.
Specifically, to sample a video, we sample a single static vector $\zs$ from the unit normal distribution and then interpolate the dynamic latent vector between two unit normally-distributed samples $\zd[1]$ and $\zd[2]$. For the spatial maps, we again sample $\Ss$ and $\Sd[1]$ from a unit normal distribution and then warp the $\Sd$ tensor continuously using a
homography transform parameterized by displacements of two upper corners and two points at the horizon. The direction of the homogrpahy is sampled randomly, speed was chosen to match the average speed of clouds in our dataset. The homography is flipped vertically for positions below the horizon to mimic the reflection process. To obtain $\Sd[i]$, we make a composition of $i-1$ identical transforms and then apply it to $\Sd[1]$. As we interpolate/warp the latent variables, we pass them through the trained model to obtain the smooth videos (\fig{synth} and \textbf{Supplementary video}). Note that our models requires no image-specific user input.

\subsection{Animating Real Scenery Images with Our Model}

\textbf{Inference.} To animate a given scenery image $I$, we find (infer) a set of latent variables that produce such image within the generator. Following \cite{Abdal2019Image2StyleGANHT}, we look for extended latents $\Wall$ and $\Sall$, so that $\G(\Wall,\Sall)\approx I$, but our procedure is different from theirs. After that, we apply the same procedure as described above to animate the given image. 

The latent space of our generator is highly redundant, and to obtain good animation, we have to ensure that the latent variables come roughly from the same distribution as during the training of the model (most important, $\Wall$ should belong to the output manifold of $\M$). Without such prior, the latent variables that generate good reconstruction might still result in implausible animation (or lack of it). We therefore perform inference using the following three-step procedure:
\begin{enumerate}
    \item \textbf{Step 1}: predicting a set of style vectors $\Wall'$ using a feedforward \textit{encoder} network $\E$ \cite{Zhu16}. The encoder has ResNet-152~\cite{he2016deep} architecture and is trained on 200000 synthetic images with mean absolute error loss.  $\Wall$ is predicted by two-layer perceptron with ReLU from the concatenation of features from several levels of ResNet, aggregated by global average pooling.
    
    \item \textbf{Step 2}: starting from $\Wall'$ and zero $\Sall$, we optimize all latents to improve reconstruction error. In addition, we penalize the deviation of $\Wall$ from the predicted $\Wall'$ (with coefficient $0.01$) and the deviation of $\Sall$ from zero (by reducing learning rate). We optimize for up to 500 steps with Adam~\cite{kingmaadam} and large initial learning rate (0.1), which is halved each time the loss does not improve for 20 iterations. A variant of our method that we evaluate separately, uses a binary segmentation mask obtained by ADE20k-pretrained~\cite{zhou2018semantic} segmentation network\footnote{CSAIL-Vision: \url{https://github.com/CSAILVision/semantic-segmentation-pytorch}}. The mask identifies dynamic (sky+water) and remaining (static) parts of the scene. In this variant, $\Ss$ (respectively $\Sd$) are kept at zero for dynamic (respectively, static) parts of the image. 
    
    \item \textbf{Step 3}: freezing latents and fine-tuning the weights of $\G$ to further drive down the reconstruction error \cite{Bau19,zakharov2019few}. The step is needed since even after optimization, the gap between the reconstruction and the input image remains. During this fine-tuning, we minimize the combination of the per-pixel mean absolute error and the perceptual loss~\cite{Johnson16}, with much larger (10$\times$) weight for the latter. We do 500 steps with ADAM and $lr=0.001$.
\end{enumerate}

\begin{figure}
    \centering
    {\renewcommand{\arraystretch}{0.5}
    \begin{tabular}{c c|c c c c c c c}
        & Input &&&&&& \\
        \rotatebox[origin=lU]{90}{$\quad$EOIF} &
            \includegraphics[height=4.9em]{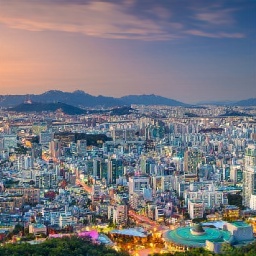} & 
            \includegraphics[height=4.9em]{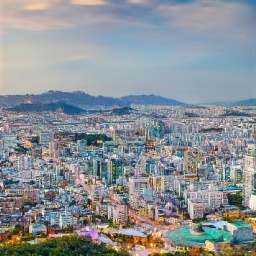} & 
            \includegraphics[height=4.9em]{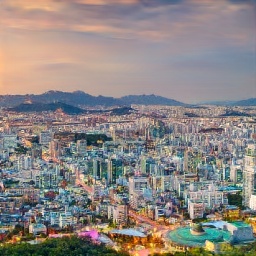} & 
            \includegraphics[height=4.9em]{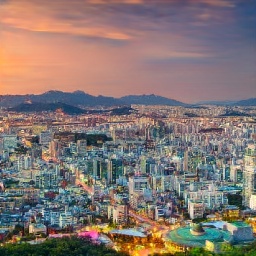} & 
            \includegraphics[height=4.9em]{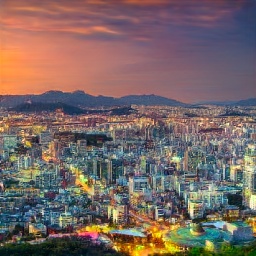} & 
            \includegraphics[height=4.9em]{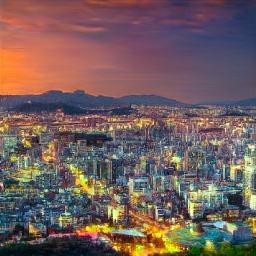} & 
            \includegraphics[height=4.9em]{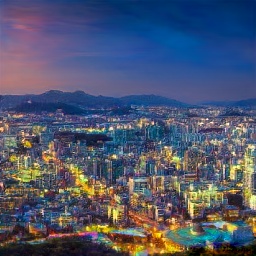}
        \\
        \rotatebox[origin=lU]{90}{$\quad$EOIF} &
            \includegraphics[height=4.9em]{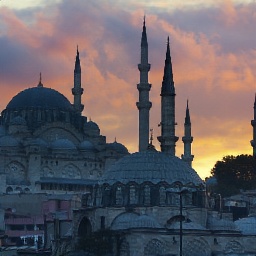} & 
            \includegraphics[height=4.9em]{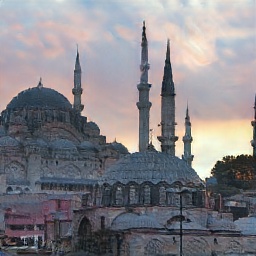} & 
            \includegraphics[height=4.9em]{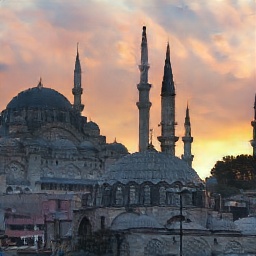} & 
            \includegraphics[height=4.9em]{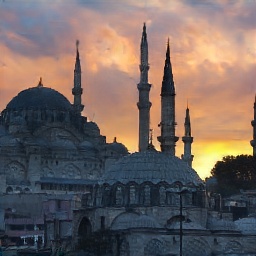} & 
            \includegraphics[height=4.9em]{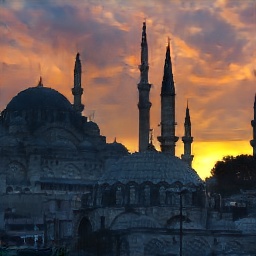} & 
            \includegraphics[height=4.9em]{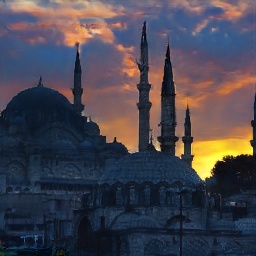} & 
            \includegraphics[height=4.9em]{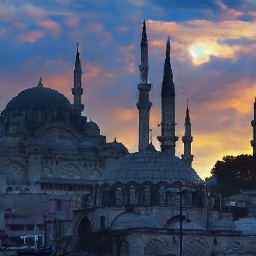}
        \\
        \rotatebox[origin=lU]{90}{$\quad$EOIF} &
            \includegraphics[height=4.9em]{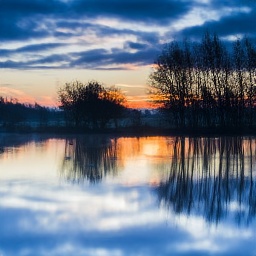} &
            \includegraphics[height=4.9em]{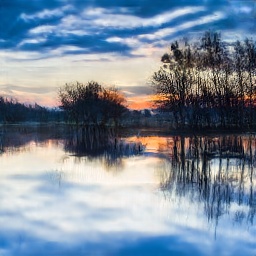} &
            \includegraphics[height=4.9em]{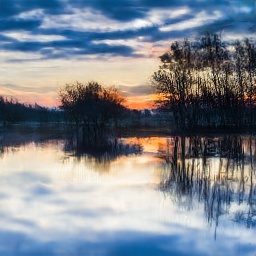} &
            \includegraphics[height=4.9em]{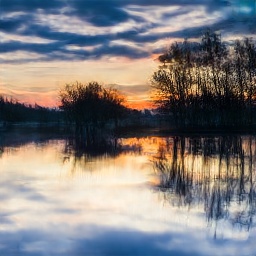} &
            \includegraphics[height=4.9em]{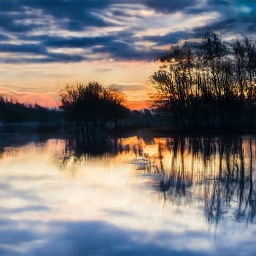} &
            \includegraphics[height=4.9em]{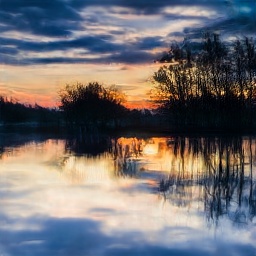} &
            \includegraphics[height=4.9em]{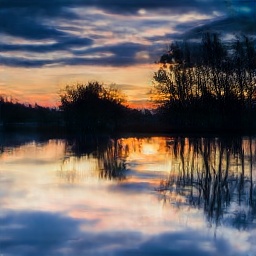}
        \\
        \rotatebox[origin=lU]{90}{$\quad$EOIFS} &
            \includegraphics[height=4.9em]{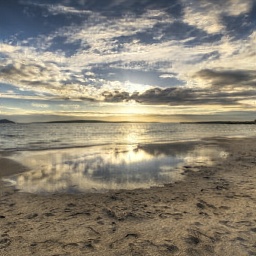} &
            \includegraphics[height=4.9em]{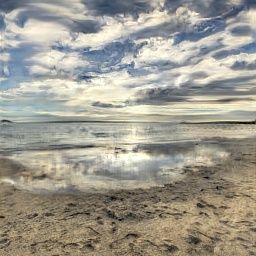} &
            \includegraphics[height=4.9em]{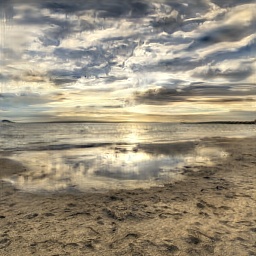} &
            \includegraphics[height=4.9em]{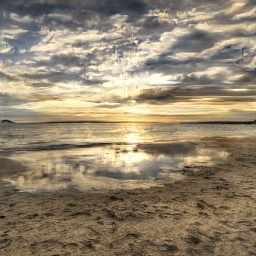} &
            \includegraphics[height=4.9em]{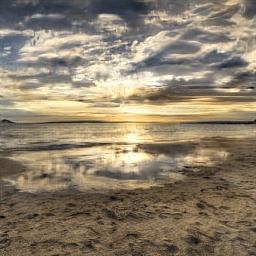} &
            \includegraphics[height=4.9em]{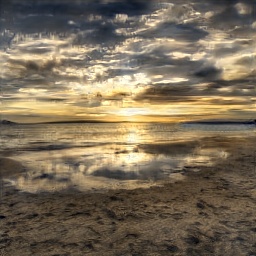} &
            \includegraphics[height=4.9em]{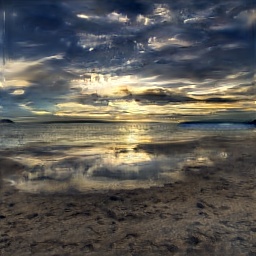}
        \\
        \rotatebox[origin=lU]{90}{$\quad$EOIFS} &
            \includegraphics[height=4.9em]{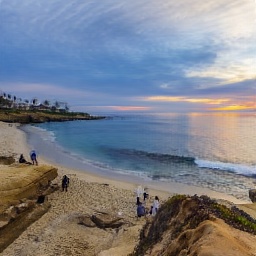} &
            \includegraphics[height=4.9em]{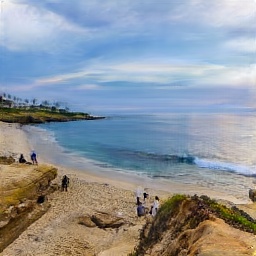} &
            \includegraphics[height=4.9em]{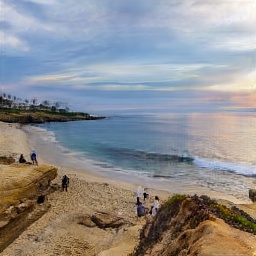} &
            \includegraphics[height=4.9em]{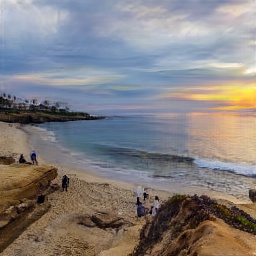} &
            \includegraphics[height=4.9em]{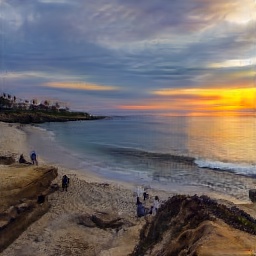} &
            \includegraphics[height=4.9em]{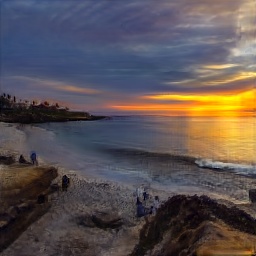} &
            \includegraphics[height=4.9em]{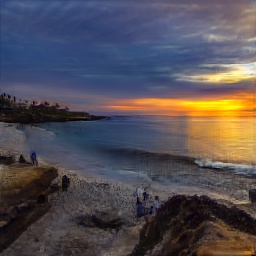}
        \\
    \end{tabular}
    }

    \caption{Examples of real images animated with our model. Each row shows a sequence of frames from a single video. Each frame is $256 \times 256$ (please zoom in for details). Clouds, reflections and waves move and change their shape naturally; time of day also changes. More examples are available in the \textbf{Supplementary video}.}
    \label{fig:real-animations}
\end{figure}

Please refer to \fig{inference-comparison} and \textit{Appendix} for examples of qualitative effects of fine tuning. We also evaluate our inference pipeline quantitatively (see Section~\ref{sect:experiments}).

\textbf{Lighting manipulation.} During training of the model, $\M$ is used to map $\zall$ to $\w$. We resample $\zd$ in order to take into account variations of lighting, weather changes, etc.\ and to have $\zs$ describe only static attributes (land, buildings, horizon shape, etc.). To change lighting in a real image, one has to change $\zd$ and then use MLP to obtain new styles $\Wall$. Our inference procedure, however, outputs $\Wall$ and we have found it very difficult to invert $\M$ and obtain $\zall = \M^{-1}(\w)$.

To tackle this problem, we train a separate neural network, $\A$, to approximate local dynamics of $\M$. Let $\w_a = \M(\zs_a, \zd_a)$ and $\w_b = \M(\zs_b, \zd_b)$, we optimize $\A$ as follows:  $\A(\w_a, \zd_b, c) \approx \M(\zs_a, \zd_a \sqrt{1 - c} + \zd_b \sqrt{c}),$ where $c \sim Uniform(0, 1)$ is coefficient of interpolation between $\w_a$ and $\w_b$. Thus, $c = 0$ corresponds to $\zd_a$, so $\A(\w_a, \zd_b, 0) \approx \w_a$; $c = 1$ corresponds to $\zd_b$, so $\A(\w_a, \zd_b, 1) \approx \w_b$.

We implement this by the combination of L1-loss $L^{\A}_{Abs} = \left| \w_b - \A(\cdot) \right|$ and relative direction loss $L^{\A}_{Rel} = 1 - \cos \left( \w_b - \w_a, \A(\cdot) - \w_a \right)$. The total optimization criterion is $L^{\A} = L^{\A}_{Abs} + 0.1 L^{\A}_{Rel}$. We train $\A$ with ADAM~\cite{kingmaadam} until convergence. At test time, the network $\A$ allows us to sample a random target $\zd_b$ and update $\Wall$ towards it by increasing the interpolation coefficient $c$ as the animation progresses. Please refer to \fig{real-animations} and \textbf{Supplementary Video} for examples of animations with our full pipeline.

\textbf{Super Resolution (SR).} As our models are trained at medium resolution (e.g.\ 256$\times$256), we aim to bring fine details from the given image that we need to animate through a separate super-resolution procedure. The main idea of our super resolution approach is to borrow as much as possible from the original high-res image (which is downsampled for animation via $\G$). To achieve that, we super-resolve the animation and blend it with the original image using a standard image superresolution approach. We use ESRGANx4~\cite{wang2018esrgan} trained on a dedicated dataset that is created as follows. To obtain the (hi-res, low-res) pair, we take a frame $I$ from our video dataset as a hi-res image, we downsample it and run the first two steps of inference and obtain an (imperfect) low-res image. Thus, the network is trained on a more complex task than superresolution.

\begin{figure}
    \centering

        \begin{tabular}{c c c c c}
            Input & $\G_1$ & $\G_2$   & $SR_1$  & $SR_2$ \\
            \includegraphics[height=7em]{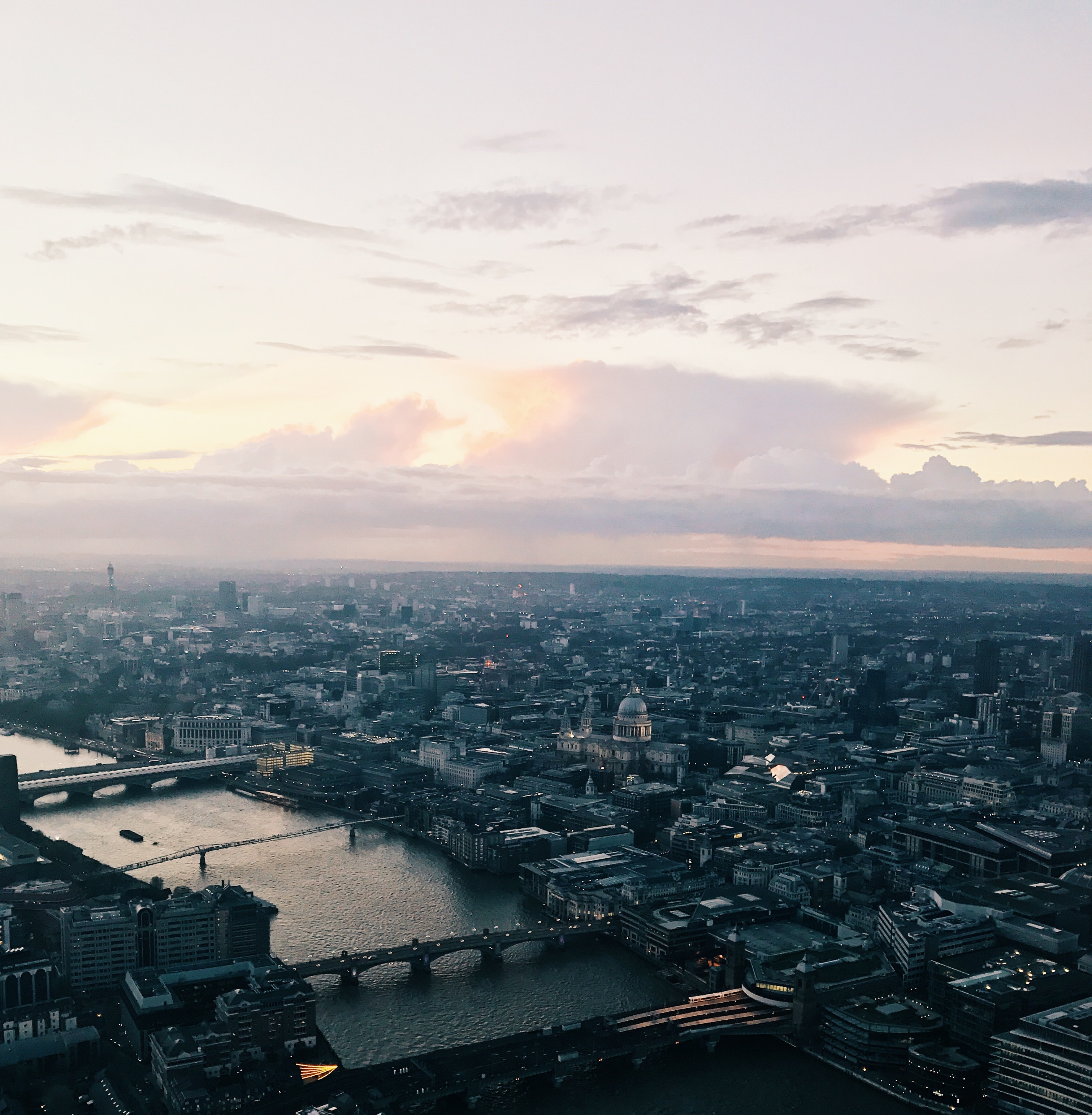} &
            \includegraphics[height=7em]{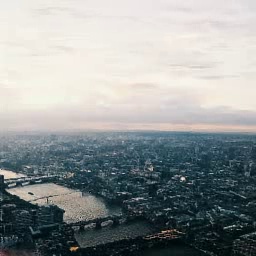} &
            \includegraphics[height=7em]{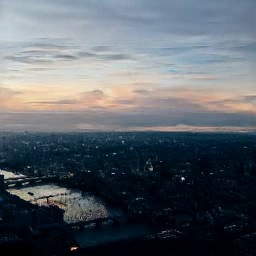} &
            \includegraphics[height=7em]{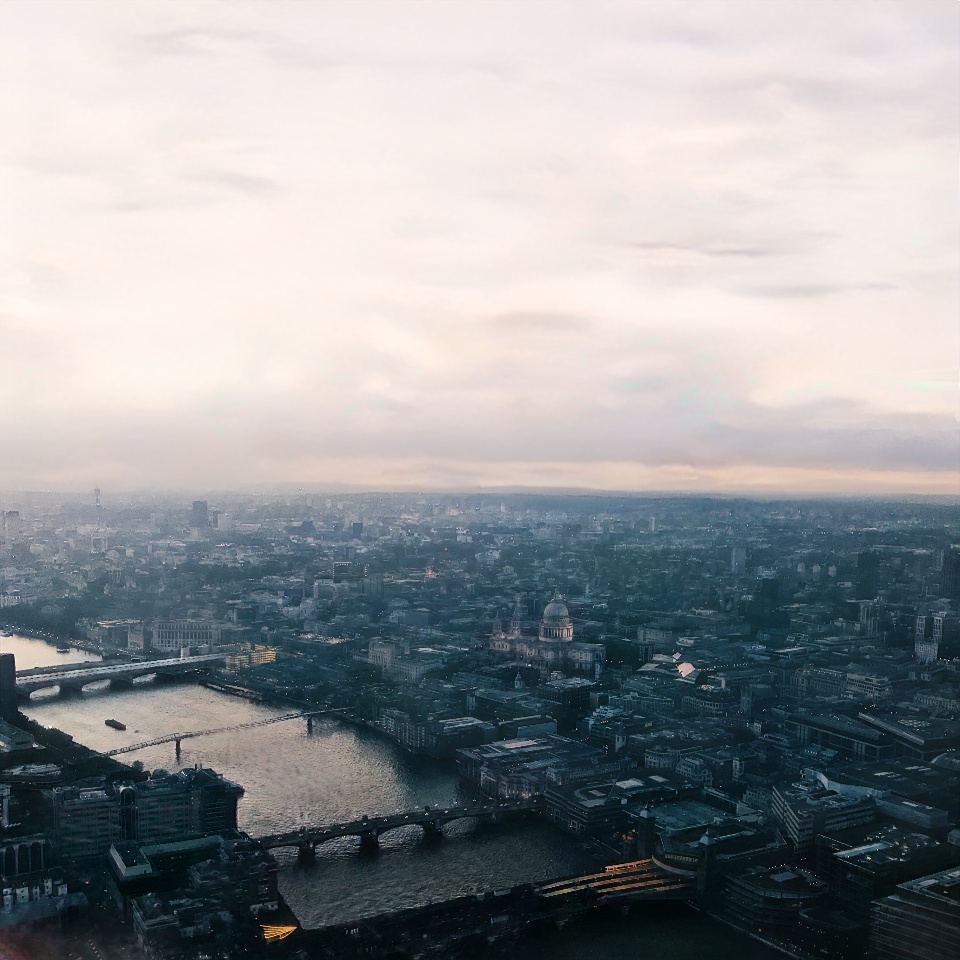} &
            \includegraphics[height=7em]{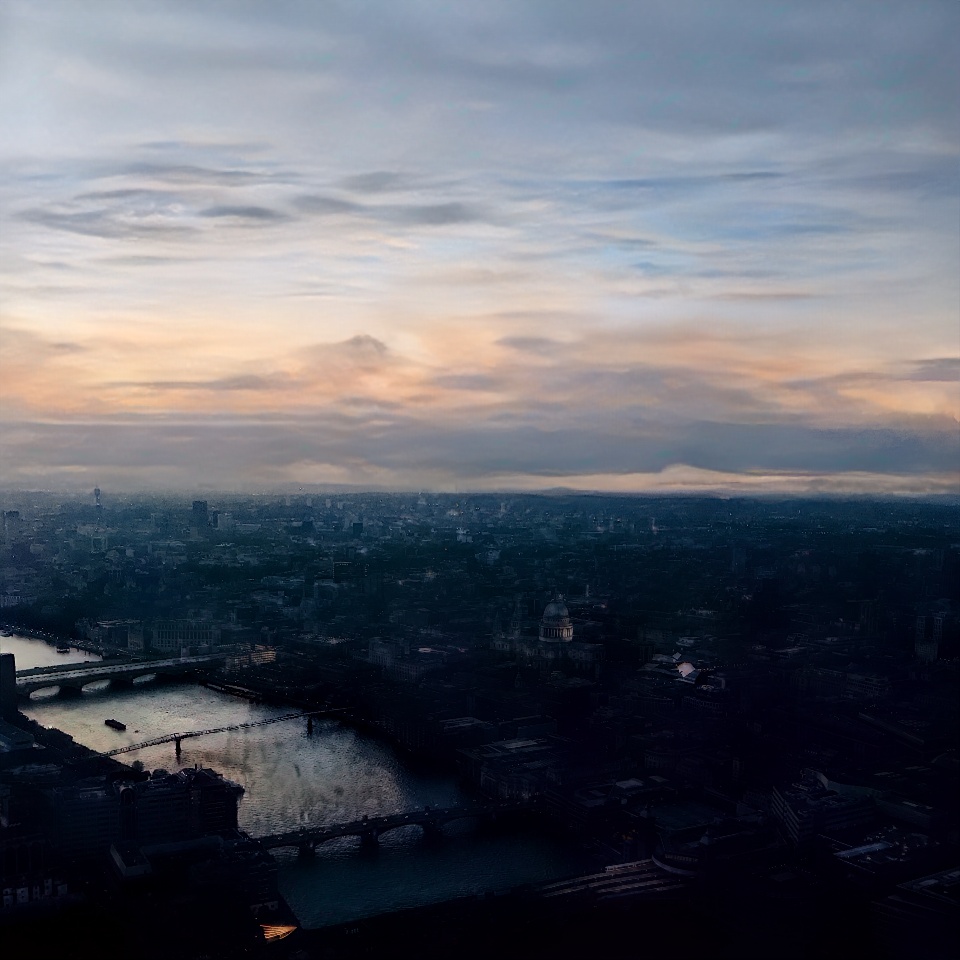} \\
            \includegraphics[height=7em]{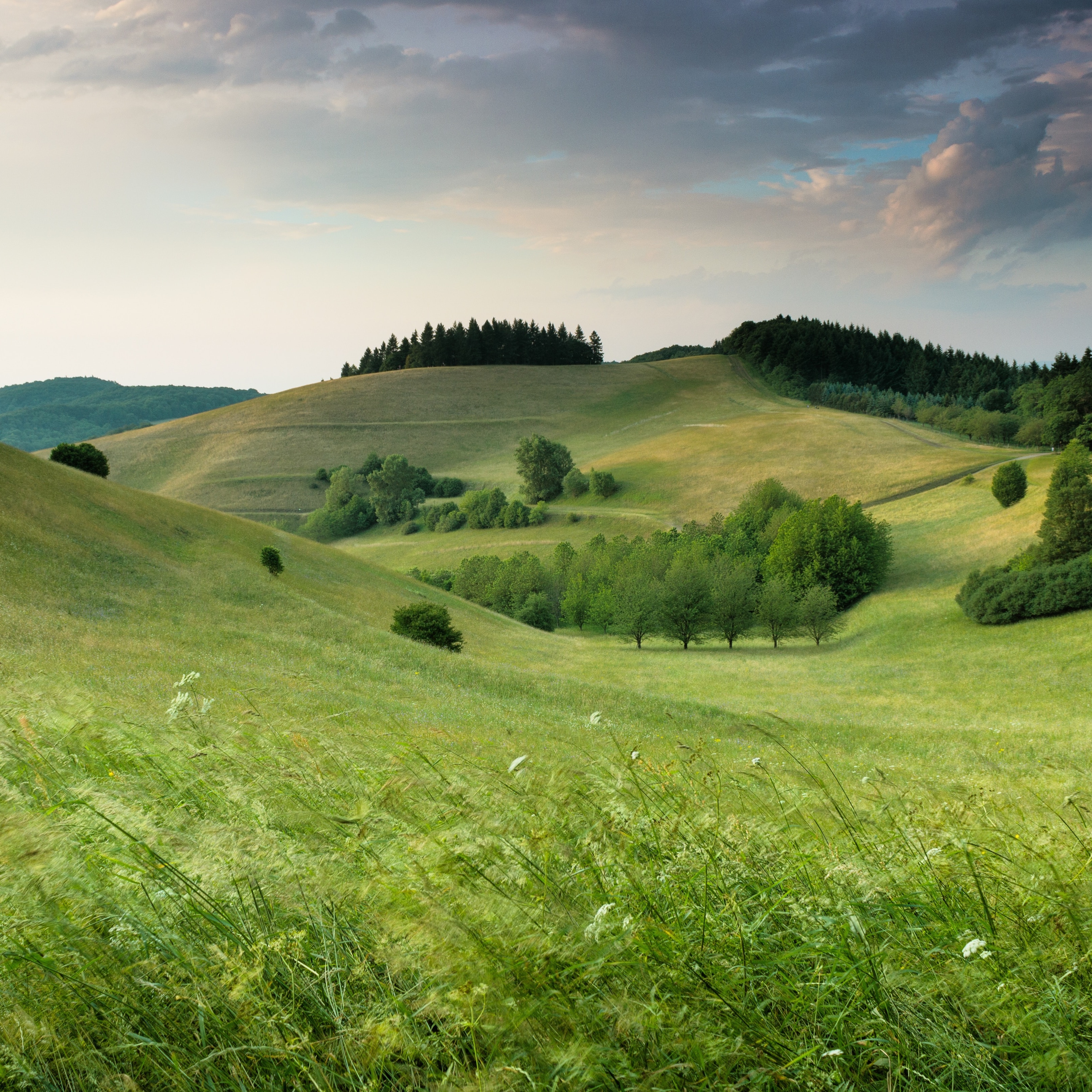} &
            \includegraphics[height=7em]{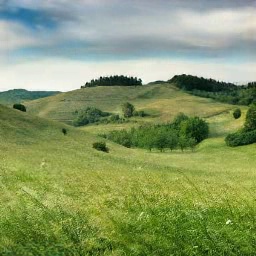} &
            \includegraphics[height=7em]{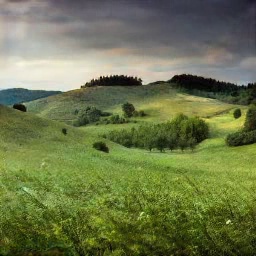} &
            \includegraphics[height=7em]{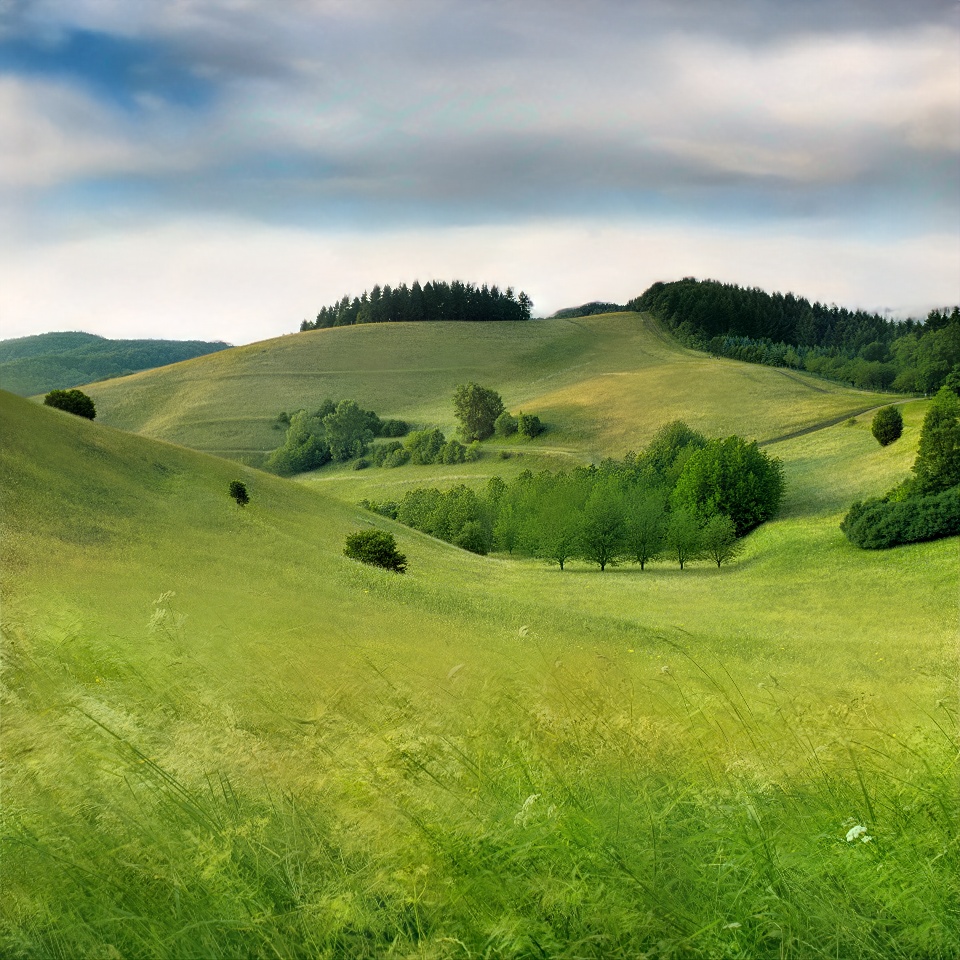} &
            \includegraphics[height=7em]{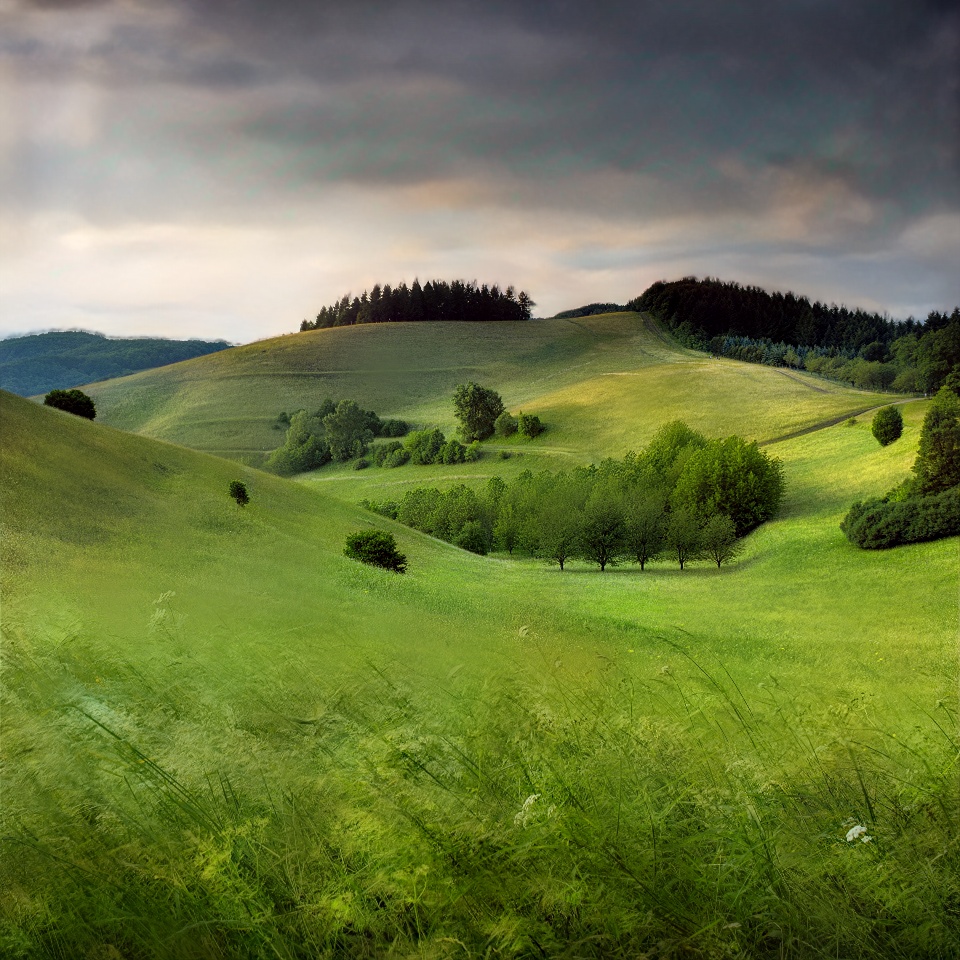} \\
            \includegraphics[height=7em]{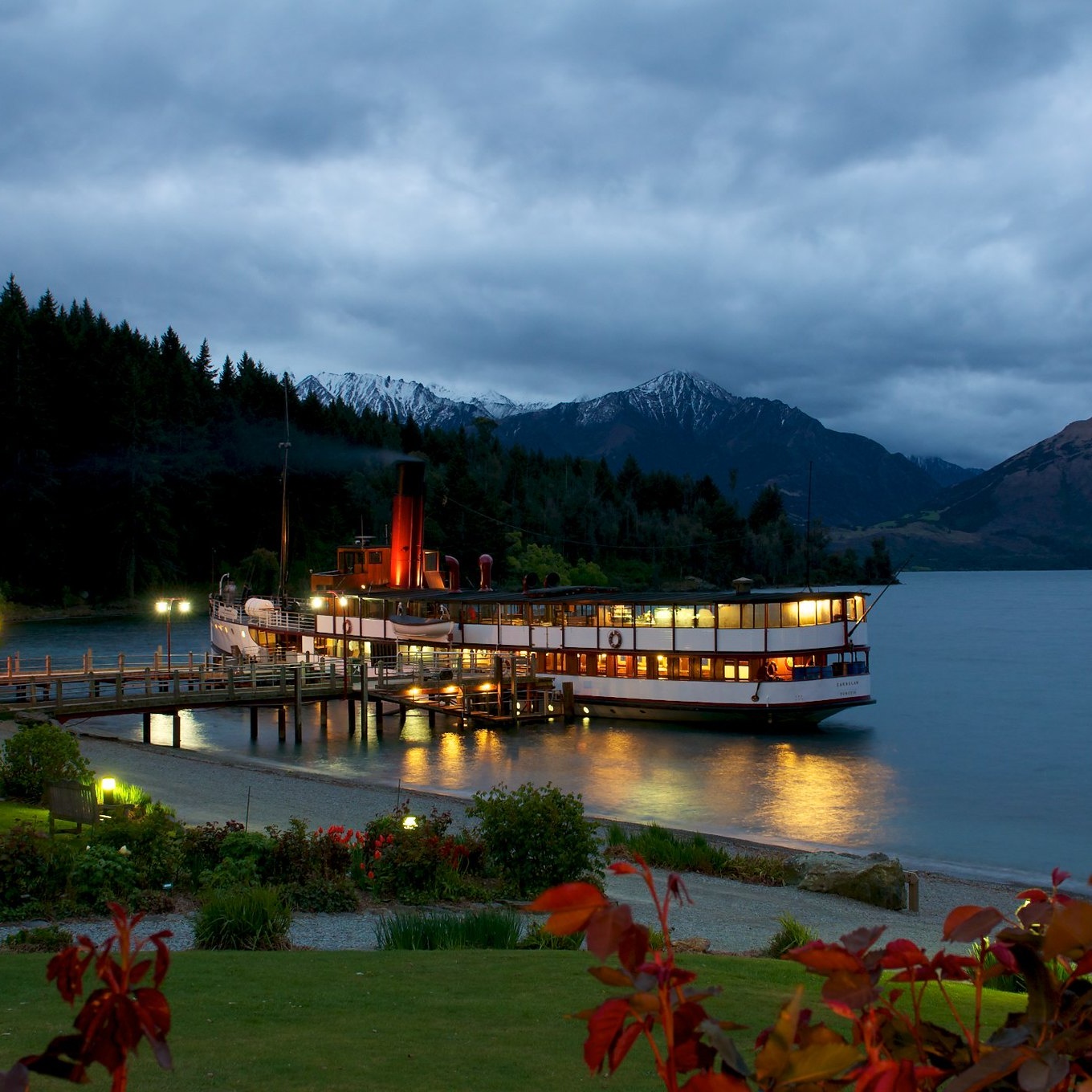} &
            \includegraphics[height=7em]{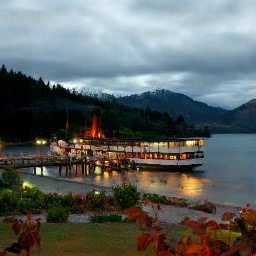} &
            \includegraphics[height=7em]{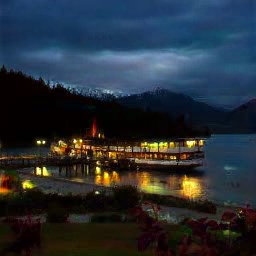} &
            \includegraphics[height=7em]{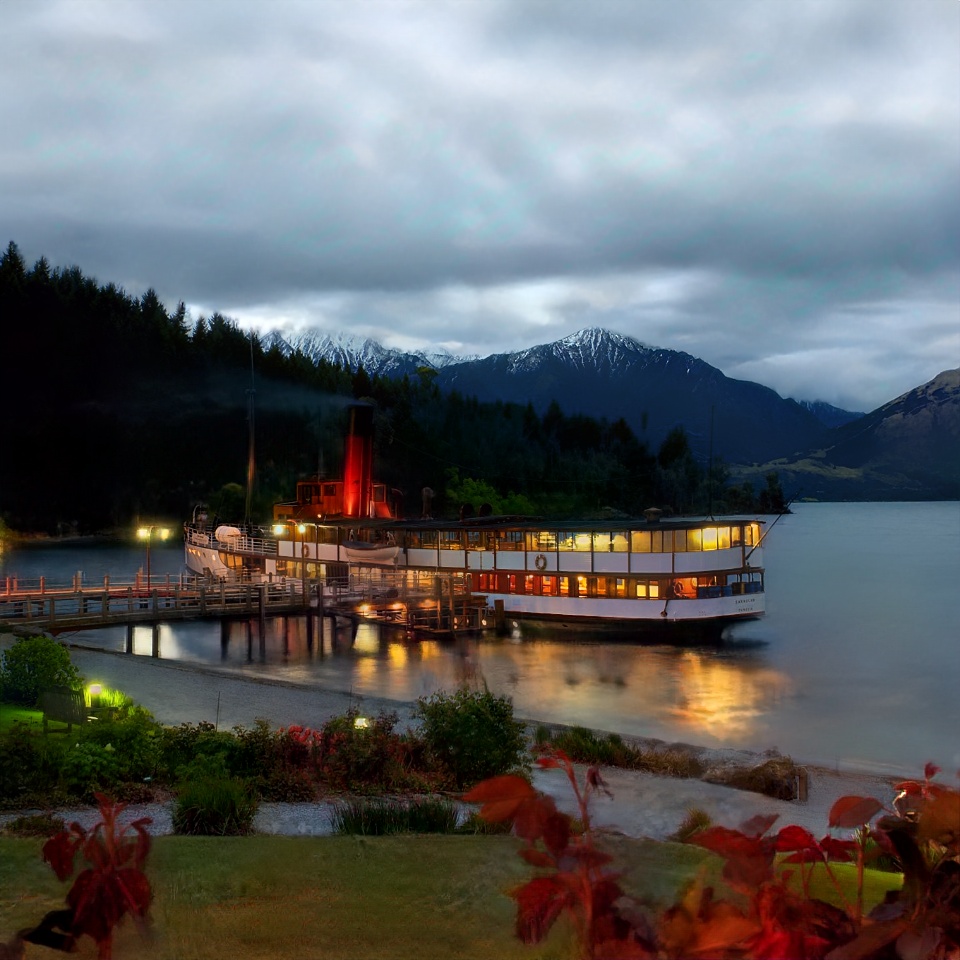} &
            \includegraphics[height=7em]{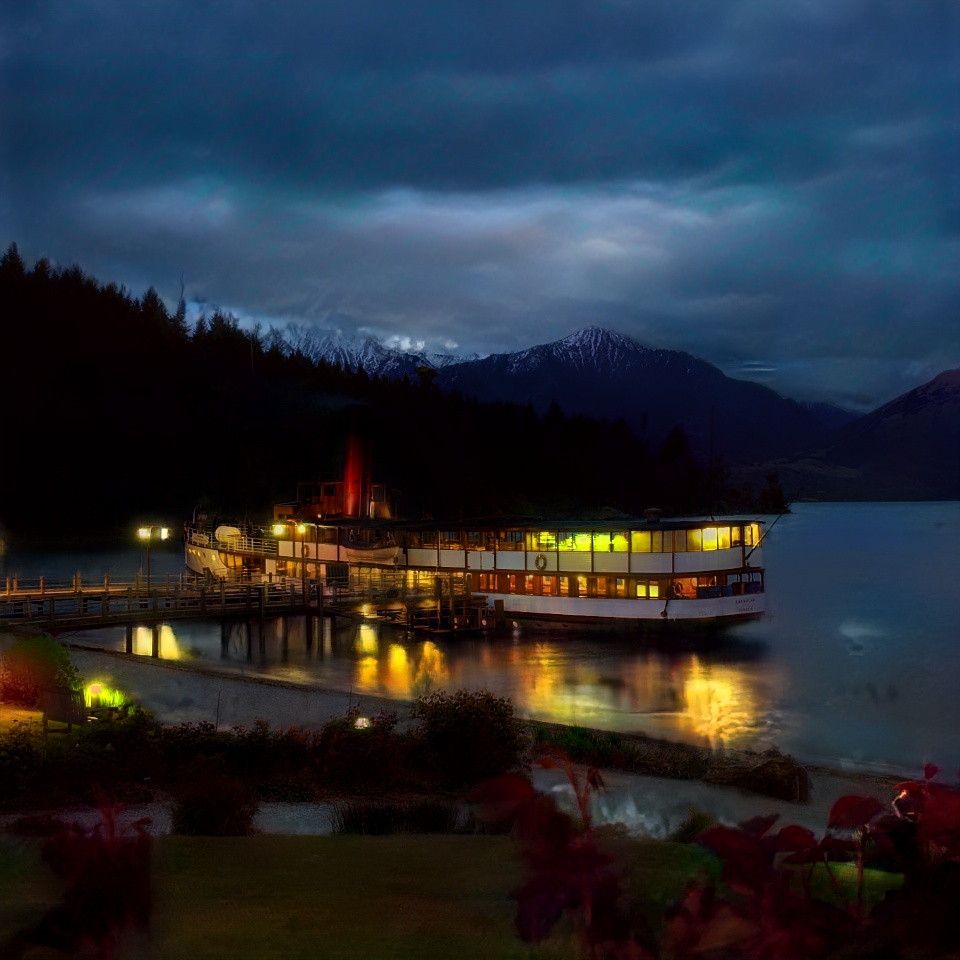} \\
            \includegraphics[height=7em]{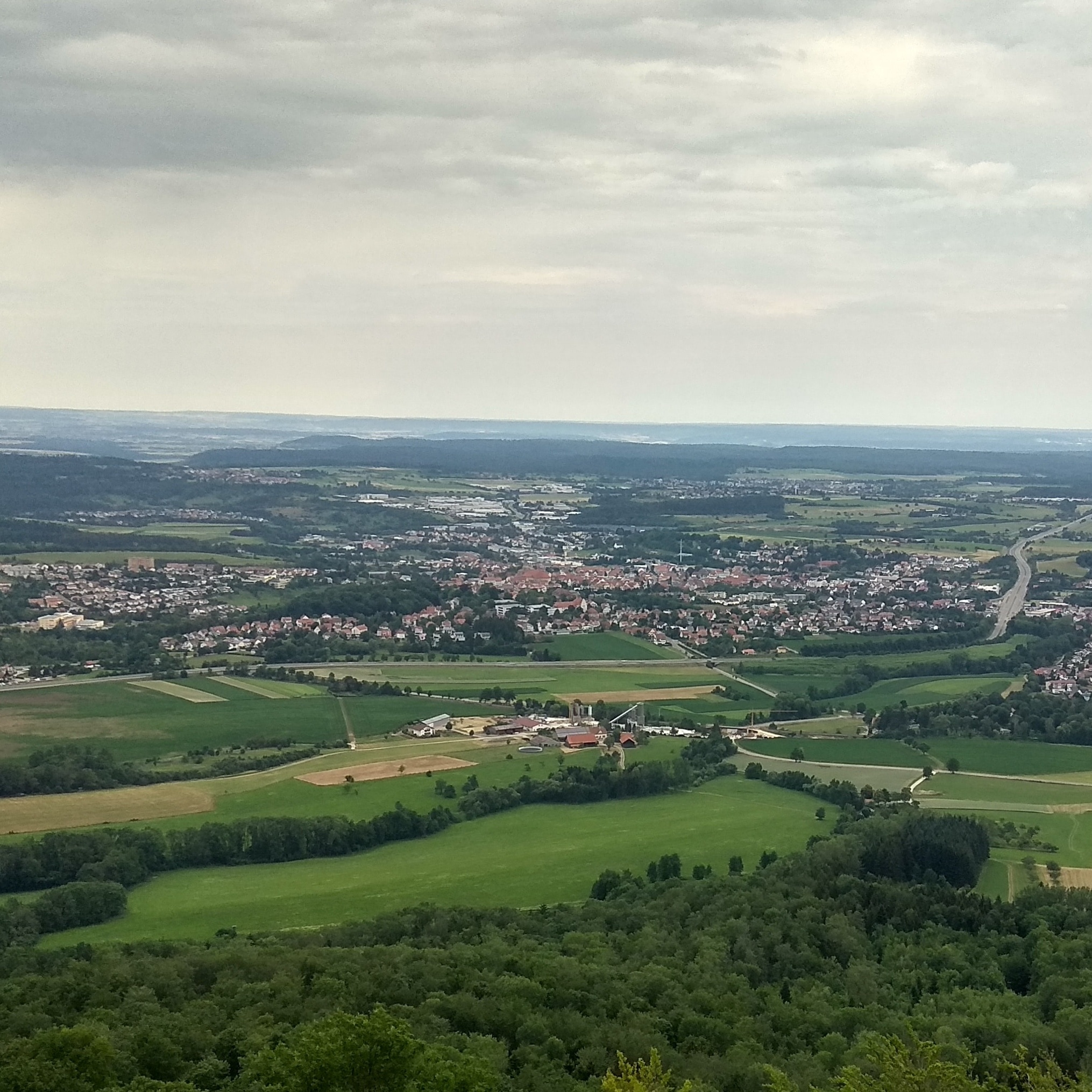} &
            \includegraphics[height=7em]{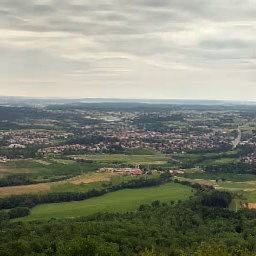} &
            \includegraphics[height=7em]{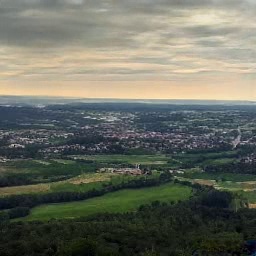} &
            \includegraphics[height=7em]{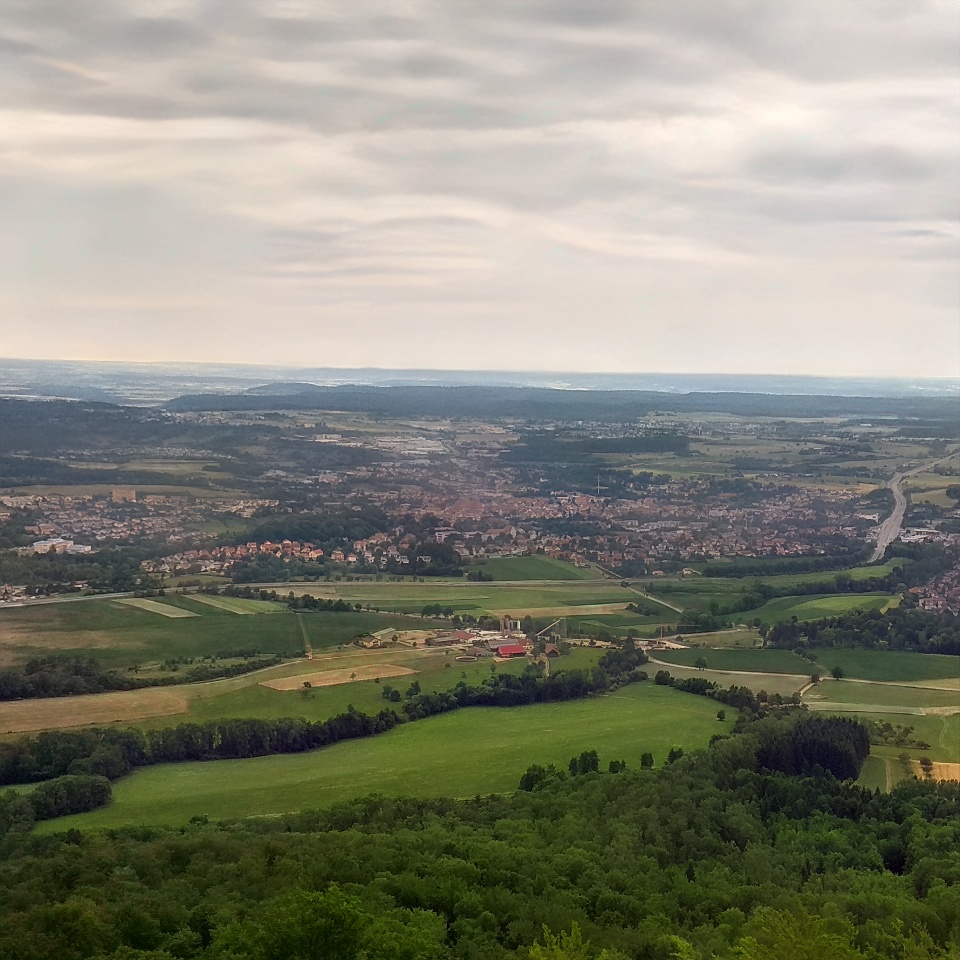} &
            \includegraphics[height=7em]{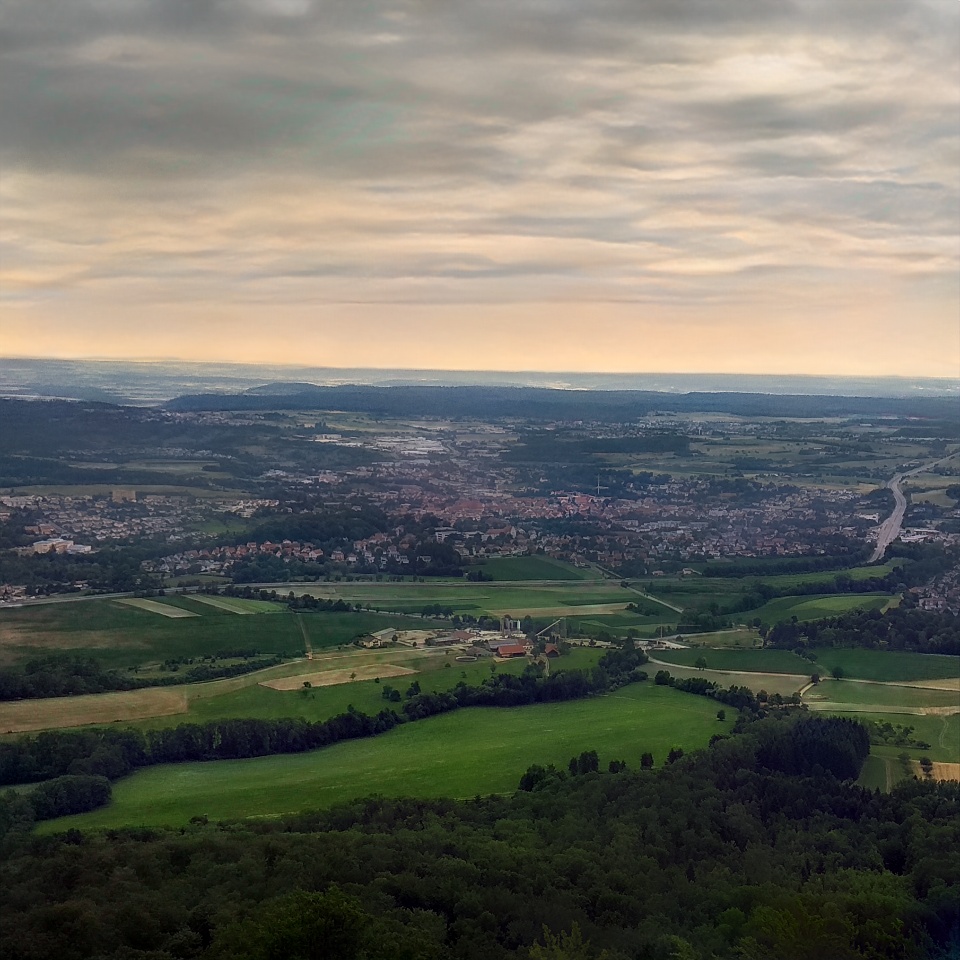} \\
            \includegraphics[height=7em]{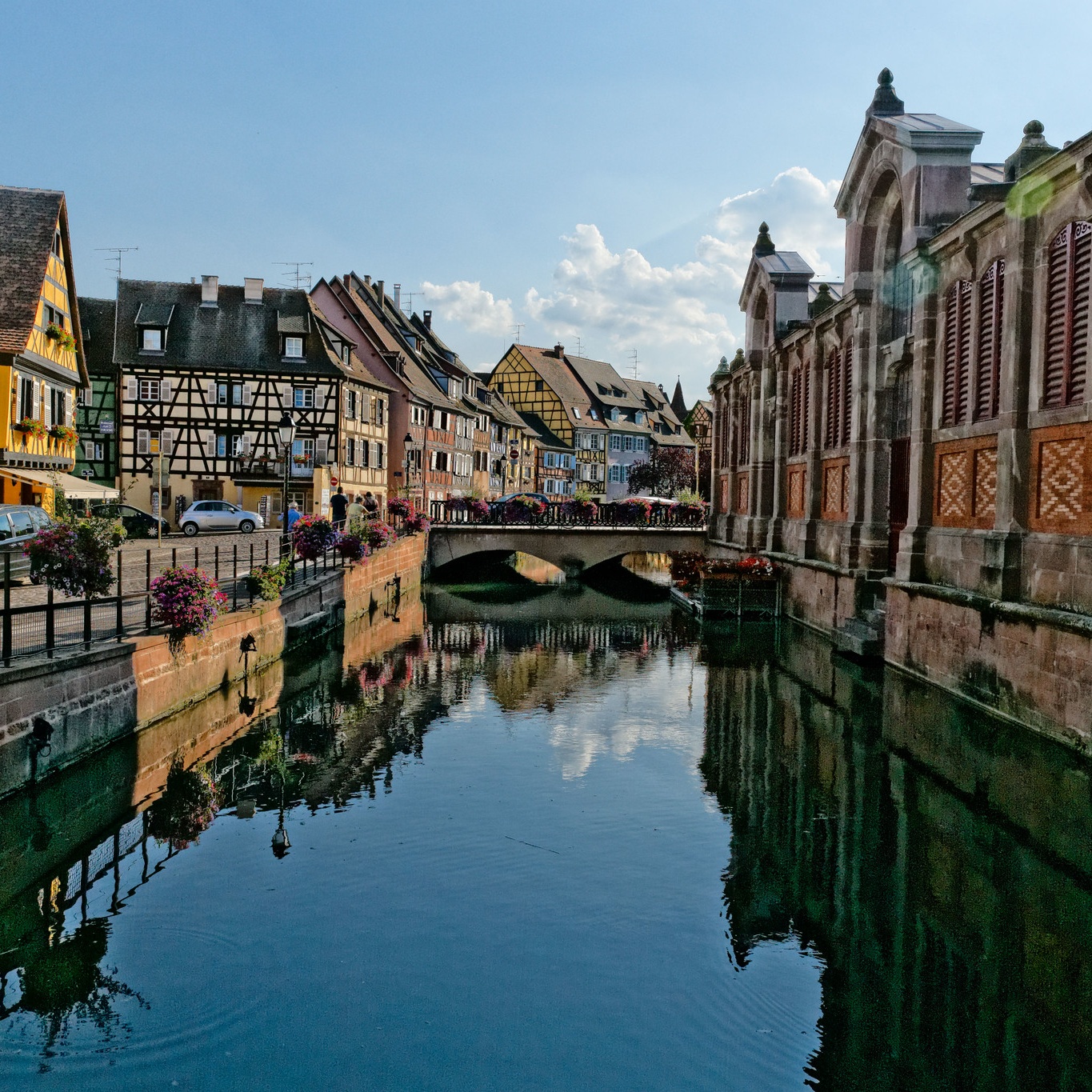} &
            \includegraphics[height=7em]{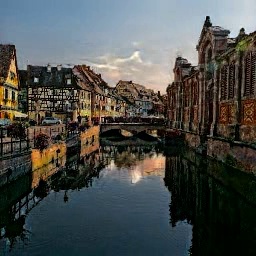} &
            \includegraphics[height=7em]{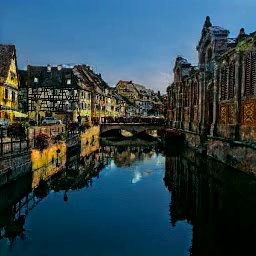} &
            \includegraphics[height=7em]{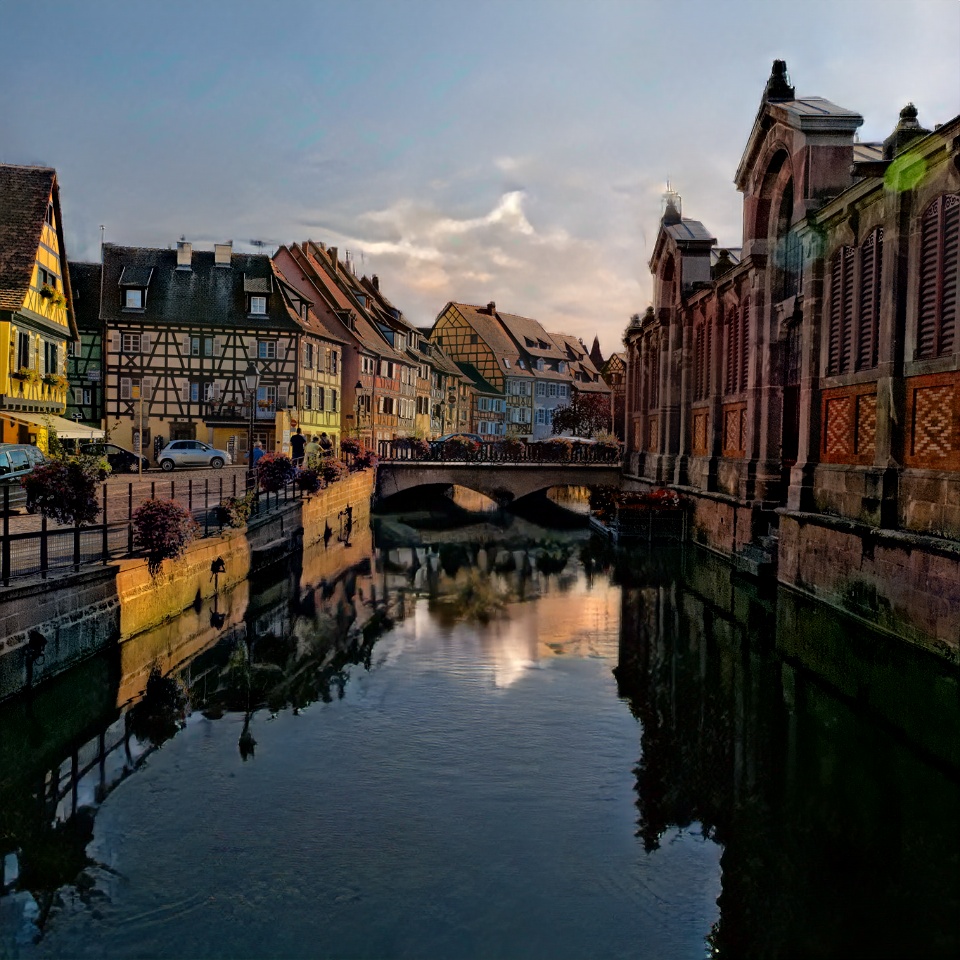} &
            \includegraphics[height=7em]{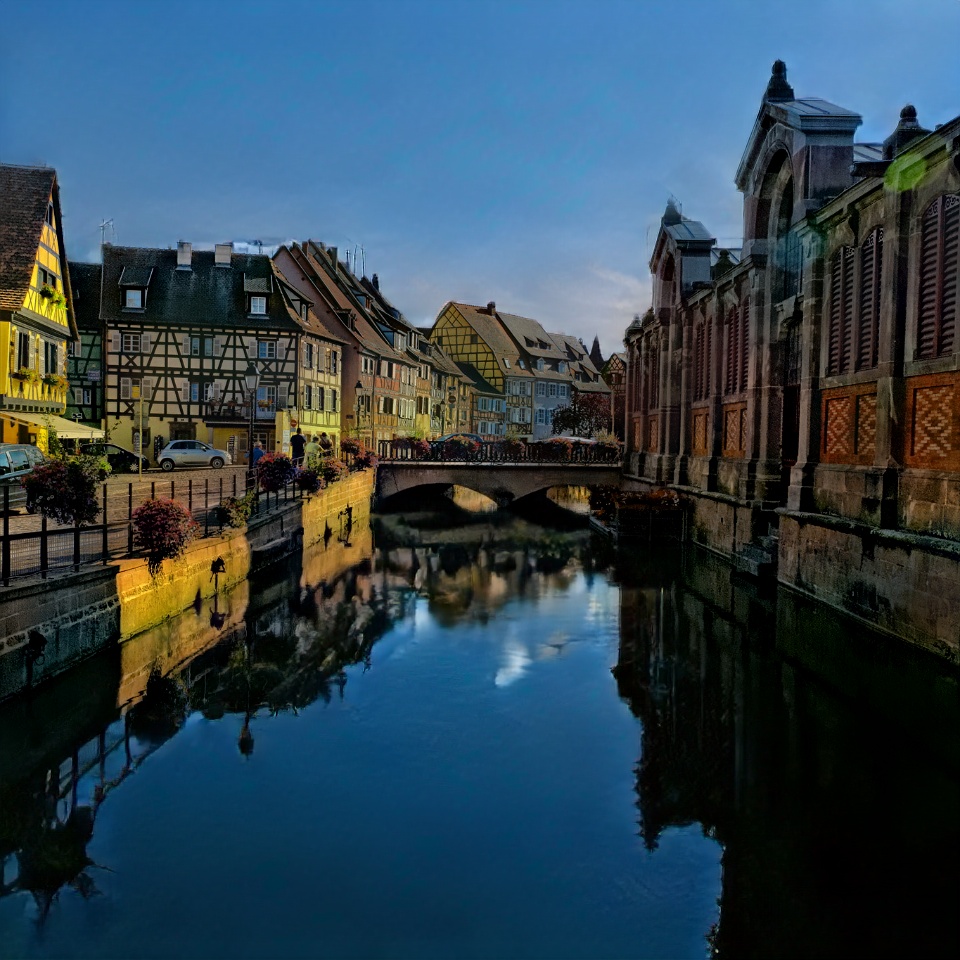} \\
    \end{tabular}

    \caption{Examples of super-resolution ($SR$) applied to the output of our generator ($\G$) given input image (Input). The inputs and SR are at $1024\times{}1024$ resolution, while the low-res images are at $256\times{}256$ resolution. Zoom-in recommended.}
    \label{fig:sr-examples}
\end{figure}

After obtaining the super-resolved video, we transfer dynamic parts (sky and water) from it to the final result. The static parts are obtained by running the guided filter~\cite{huikai2018gudied} on the super-resolved frames while using the input high-res image as a guide. Such procedure effectively transfers high-res details from the input, while retaining the lighting change induced by lighting manipulation (\fig{sr-examples}).

\section{Experiments}
\label{sect:experiments}

We evaluate our method both quantitatively and qualitatively (via user study) on synthetic and real images separately. Evaluation on synthetic images (\textit{generation}) aims on quantifying impact of major design choices of $\G$ itself (without encoding and super-resolution). Evaluation on real images (\textit{animation}) aims on comparison with previous single-image animation methods, including Animating Landscape (\textit{AL})~\cite{Endo2019AnimatingLS}, SinGAN (\textit{SG})~\cite{Shaham2019SinGANLA} and Two-Stream Networks (\textit{TS})~\cite{Tesfaldet2017TwoStreamCN}.
The Animating Landscape system is based on learnable warping and is trained on more than a thousand time-lapse videos from \cite{zhu2017unpaired,Xiong_2018_CVPR}.
The SinGAN method creates a hierarchical model of image content based on the input model alone. It therefore has an advantage of not needing an external dataset, though, as a downside, it requires considerable time to fit a new image. Two-Stream Networks~\cite{Tesfaldet2017TwoStreamCN} create animated textures given a static texture image and a short clip (an example of motion) via optimization of video tensor. We also tried a to include two more baselines, i.e. linear dynamic systems \cite{yan2004} and Seg2Vid~\cite{pan2019video}, but with former we got very poor quality and the latter failed to converge on our data, so we did not proceed with full comparison. We also tried to train and finetune \textit{AL} on our video dataset (which is significantly smaller than that from \textit{AL} paper), with little success (see supp.mat.).

We estimate quality through three different aspects: \textit{individual image quality}; \textit{static consistency}; \textit{animation plausibility}. Individual image quality is estimated via Fréchet Inception Distance~\cite{Heusel2017GANsTB}, masked SSIM~\cite{wang2004image} and LPIPS~\cite{zhang2018unreasonable}. Static consistency evaluation aims on quantifying how good objects that must not move (e.g. buildings, mountains etc.) are preserved over time. For that purpose we calculate SSIM and LPIPS between first frame and each generated video frame (only for static parts). Perfect image quality and static consistency can be achieved by not animating anything at all. Thus, we evaluate animation plausibility via user study and Fréchet Video Distance~\cite{unterthiner2018towards}.

To generate videos using our method, we use a manually constructed set of homographies. Data-driven estimation of homographies is out scope of this work, so we have prepared 12 homographies, one for each clock position (e.g.\ the ``12h" move clouds up and towards the observer, the ``3h" moves straight to the right, etc.). Normally, these homographies resemble the average speed of clouds in our training dataset.
We increase this speed for synthetic experiments to make differences between variants of our method more obvious; we slow down animation for experiments with real images in order to approximately align our speed with that of the competitors (AL, SG and TS).

\textbf{Datasets.} Our model was trained using both videos and single images available in the Internet under Creative Commons License. For evaluation we use 69 landscape FullHD time-lapse videos published on YouTube between Dec. 28 2019 and Jan. 29 2020. 
For FID  computation, we have collected 2400 pictures from Flickr\footnote{https://flickr.com}.



\begin{figure}
    \centering
        \begin{tabular}{l|c c c c}
            \hline\hline
            Setup           & FID$\downarrow$   & SSIM$\uparrow$& LPIPS$\downarrow$ & $\Delta$R\\
            \hline
            Original StyleGAN & \textbf{48.40}   & 0.809 & \textbf{0.049} &    \\ 
            \hline
            + frame discriminator  & 55.92   & 0.846 &  0.064    & 0.13   \\
            + separate $\Ss$ and $\Sd$  &55.15  &0.854 & 0.073 & 0.01  \\
            + separate $\zs$ and $\zd$  &54.38  &0.879       & 0.065  & 0.03\\
            + crop sampling  & 56.13   & \textbf{0.884} & 0.062 & \textbf{0.06} \\
            \hline\hline
        \end{tabular}

    \caption{Results of the ablation study of our model for the task of new video generation. The column $\Delta$R in the table
    are obtained from the side-by-side user study. $\Delta$R shows the increase in frequency when assessors prefer this variant to that in previous row (+0.23 against original StyleGAN).
    }
    \label{fig:gen-ablation}
\end{figure}

\textbf{Generation}. In order to perform thorough ablation study in reasonable time, we perform all evaluations in this section at $128 \times 128$ resolution. 
To estimate static consistency, we sample 1200 pairs of images from $\G$, mask out sky and water according to segmentation mask and calculate LPIPS and SSIM between two images in a pair. In each pair the images are generated from the same $\zs,\Ss$ and different $\zd,\Sd$. For the user study we sample 100 videos 200 frames long at 30 FPS. In order to compare different ablations, the assessors were asked to select the most realistic video from a pair shown side-by-side. Each assessor is limited to evaluate no more than three pages with four tasks on each and has five minutes to complete each page. In our user study we showed each pair to five assessors. The ablation study results (\fig{gen-ablation}) reveal that the original StyleGAN generates the most high-fidelity images, but fails to preserve details of static objects. LPIPS is more tolerant to motion until the ``texture type" changes dramatically. Thus, despite LPIPS and FID achieving the best values for the original StyleGAN, it actually does not preserve static objects (see \textbf{Supplementary Video}). Our modifications allow to keep a similar level of the FID value, but gradually improve static consistency and animation plausibility. 

\textbf{Real image animation.} Experiments in this section are performed at $256 \times 256$ resolution. To calculate quantitative and qualitative metrics, we took the first frames $I_0$ of the test videos, encoded and animated them with our method. Denote the $n$-th frames of real and generated videos as $I_n$ and $\widehat{I}_n$ respectively. With our method, for each input image we generate five variants with homographies randomly sampled from the predefined set. For AL we generate five videos for each input image with randomly sampled motion, as described in the original paper. For all quantitative evaluations we do not apply style transfer in AL and $\Wall$ manipulation in our method. We evaluate two variants of AL: with (AL) and without first-to-last interpolation (AL$_\text{noint}$), which stabilizes image quality, but makes long movements impossible. We use the official implementation\footnote{https://github.com/endo-yuki-t/Animating-Landscape} of Animating Landscape~\cite{Endo2019AnimatingLS} provided by authors.
We use pretrained AL model; we also evaluate finetuned AL model
and found that most metrics degraded, while the training loss continued to improve. This can be attributed to the fact that the video dataset used in AL is bigger than ours; both include the public part of data from \cite{Xiong_2018_CVPR}. Both datasets are just youtube landscape videos and seem to be equally close to the validation (we are not aware of any biases). Also, our dataset contains videos with very different motion speed, and neither text of AL nor its code contains details regarding video speed equalization.
All images are animated in original resolution cropped to 1:1 aspect ratio via center crop, then bilinearly downsampled to $256\times256$ resolution.

For SG~\cite{Shaham2019SinGANLA} we used the official implementation\footnote{https://github.com/tamarott/SinGAN} and default parameters. We have not noticed significant difference between multiple SG runs both in terms of quantitative metrics and visual diversity. Hence we decided not to generate similar videos many times and sampled only one video for each input image.

For TS~\cite{Tesfaldet2017TwoStreamCN} we used the official implementation\footnote{https://github.com/ryersonvisionlab/two-stream-dyntex-synth}. TS can animate only the whole image, so (1) we used semantic segmentation to extract sky; (2) transferred motion to the extracted image fragment from a random video from the validation set; (3) blended static part of the original image with the generated clip. TS is only capable of producing 12 frames due to GPU memory limitations, so we interpolated frames in order to obtain the necessary video length.

\begin{figure}
    \begin{tabular}{p{0.29\textwidth} p{0.30\textwidth} p{0.41\textwidth}}
        \parbox[c]{\textwidth}{
            \includegraphics[width=0.31\textwidth]{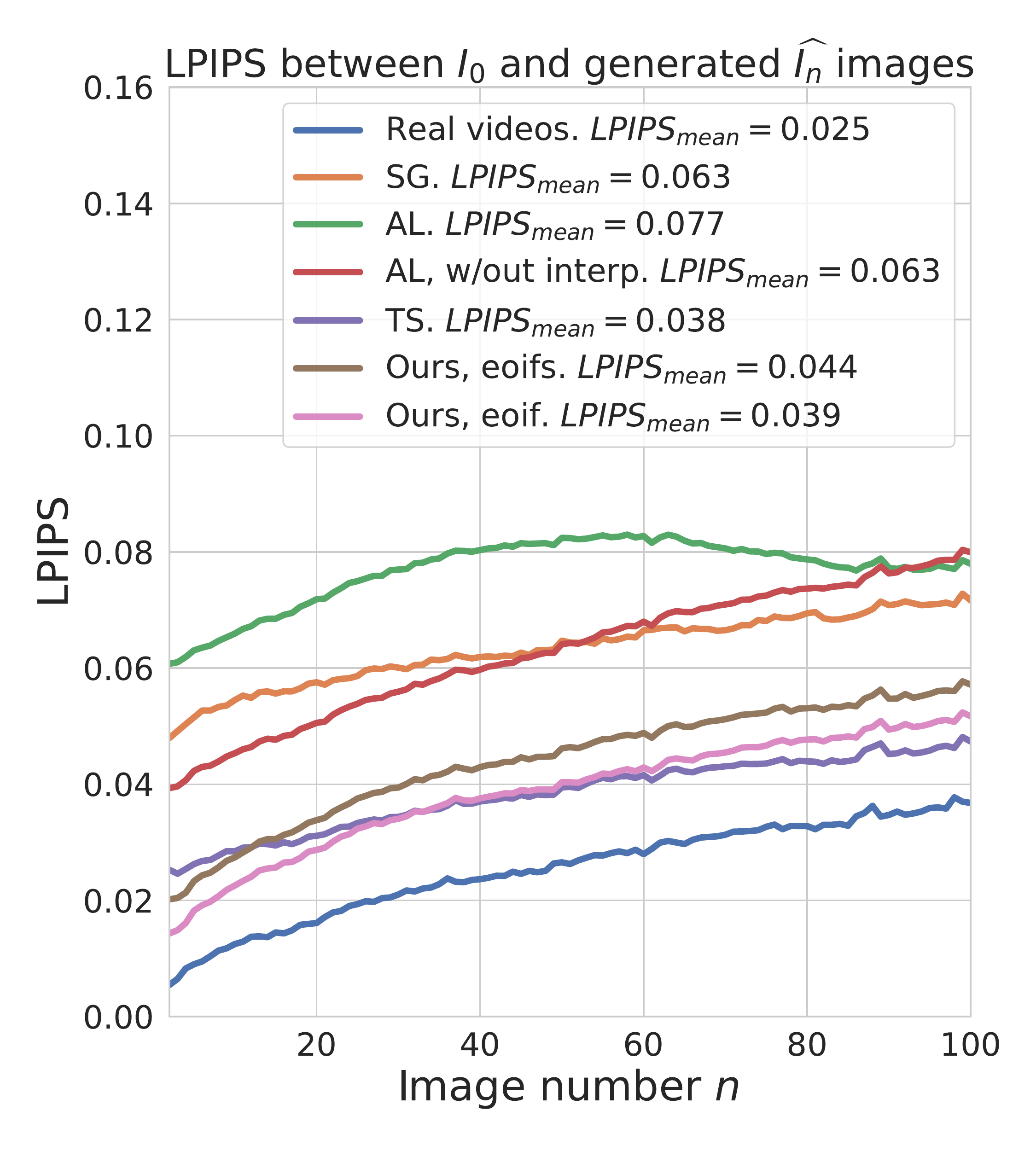}
        }
        &
        \parbox[c]{\textwidth}{
            \includegraphics[width=0.31\textwidth]{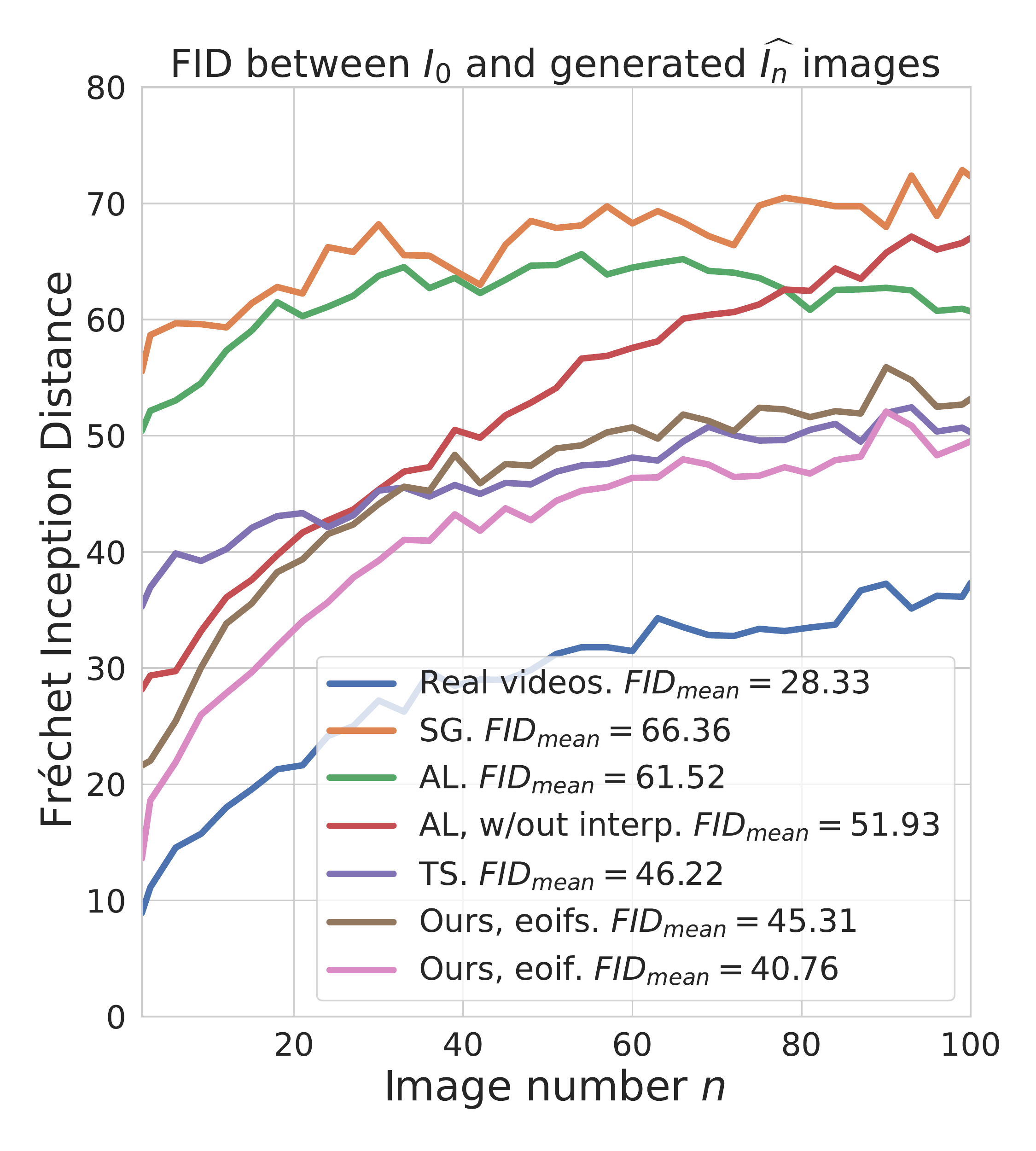}
        }
        &
        \begin{tabular}{l|c|c|c|c}
            \hline\hline
            \textbf{Name} & FVD  & LPIPS & SSIM & FID \\
            \hline\hline
            SG & 210 & 0.063 & 0.93 & 66.7 \\ 
            \hline
            AL & 275 & 0.077 & 0.91 & 61.9 \\
            AL$_\text{noint}$ & 162 & 0.063 & 0.92 & 52.4 \\
            \hline
            TS & 420 & \textbf{0.039} & \textbf{0.96} & 46.6\\
            \hline
            Ours$_\text{eoifs}$ & 161 & 0.044 & 0.94 & 45.8 \\
            Ours$_\text{eoif}$ & \textbf{149} & \textbf{0.039} & 0.95 & \textbf{41.2} \\
            \hline\hline
        \end{tabular}
    \end{tabular}
    \caption{Quantitative comparison of image quality, static consistency and motion plausibility. \textbf{Left and middle}: LPIPS$\downarrow$ and FID$\downarrow$ between $I_0$ and $\widehat{I}_n$, which mostly measure image quality and static consistency. The legend contains metrics averaged over time. As can be seen, pixel-level transformations (e.g. using predicted flows in AL) lead to faster deterioration of generated images over time, compared to our approach, especially for later frames ($n \gtrsim 50$).  \textbf{Right}: FVD$\downarrow$, LPIPS$\downarrow$, SSIM$\uparrow$ and FID$\downarrow$ between $I_n$ and $\widehat{I}_n$ averaged over time, which measure not only image quality, but also animation plausibility.
    }
    \label{fig:real-comparison}
\end{figure}

We evaluate image quality by measuring FID between the set of all first frames of real videos $I_0$ and the set of $n$-th frames of generated videos $\widehat{I}_n$. Thus, we can see how fast these two distributions diverge. Too fast divergence in terms of FID may indicate image quality degradation in time. We evaluate static consistency by measuring LPIPS between $I_0$ and $\widehat{I}_n$ with moving parts masked out according to semantic segmentation. We always predict semantic segmentation only for $I_0$. Higher LPIPS may indicate that static areas are tampered during animation (i.e.~they are erroneously moving).
We also follow the adopted practice to quantitatively measure motion similarity using Fréchet Video Distance (FVD)~\cite{unterthiner2018towards} between real and generated videos, which is averaged over motion directions. Different motion directions are obtained via sampling different homography (Ours), motion code (AL) and horizontal flipping, choosing random reference video (TS). As revealed in \fig{real-comparison}, our method preserves static details better and the speed of image quality degradation with time is slower than that of AL$_{noint}$.

\begin{figure}
    \centering
    \begin{tabular}{p{0.7\textwidth} p{0.3\textwidth}}
         \begin{tabular}{l|cc|cc}
            \hline\hline
            \textbf{Method} & \multicolumn{2}{c|}{\textbf{short}} & \multicolumn{2}{c}{\textbf{long}} \\
            \hline
            & \textbf{EOIF} & \textbf{EOIFS} & \textbf{EOIF} & \textbf{EOIFS} \\
            \hline\hline
            SG & 0.40 & 0.44 & 0.26 & 0.29
            \\ \hline
            AL (no int) & 0.46 & 0.47 & 0.37 & 0.38
            \\ \hline
            AL (+ style) & 0.18 & 0.18 & 0.11 & 0.10
            \\ \hline
            TS & 0.11 & 0.12 & 0.12 & 0.14
            \\ \hline
            Real & 0.41 & 0.44 & 0.44 & 0.45
            \\ \hline
            Ours (EOIF) & -- & 0.52 & -- & 0.52            
            \\ \hline
            Ours (EOIFS) & 0.48 & -- & 0.48 & --\\
            \hline\hline
        \end{tabular}   
        &
        \begin{tabular}{l|l}
            \hline\hline
            \textbf{}     & \textbf{FR} \\
            \hline\hline
            AL (+ style) & 0.25
            \\ \hline
            SG & 0.38
            \\ \hline
            Ours (Synth.) & 0.42
            \\ \hline
            AL (no int) &  0.54
            \\ \hline
            TS & 0.20
            \\ \hline
            Real &   0.59
            \\ \hline
            Ours (EOIFS) & 0.62 
            \\ \hline 
            Ours (EOIF) & \textbf{0.63}
            \\ 
            \hline\hline
        \end{tabular}
    \end{tabular}

    \caption{\textbf{Left}: Ratio of wins row-over-column for side-by-side settings for short (100 frames) and long (200 frames) videos. \textbf{Right}:  fooling ratio for the real/fake protocol. Note that advantage of our method becomes more evident in long videos.}
    \label{fig:real-user-study}
\end{figure}

The user study is carried out using the same real and generated videos as the ones used in quantitative evaluation. We decided to conduct two sets of user studies involving real image animation: side-by-side comparisons and real/fake questions. In the side-by-side setting, assessors are asked to select the more realistic variant of animation (from two) given the real image shown in the middle. Both videos in a pair are obtained from the same real image using different methods. In real/fake setting, assessors see only a single video and guess whether it is real or not. Each assessor was shown at most 12 questions, 5 different assessors per one question. During the study we noticed that the video speed affects user preference (slower ones are more favorable). Since we cannot control animation speed in our baselines fairly, we decided to conduct two sets of user studies: (A) with motion speed aligned to that of competitors and (B) aligned to that of real videos. Here we present only results of A setting (see supp.mat. for B setting). To sum up, the user study reveals the advantage of our method over three baselines (AL, SG, TS), especially in longer videos.

Please refer to \nameref{appendix} for more details on methods and experiments, including quantitative ablation study of inference procedure.

\section{Discussion}
\label{sect:conclusion}
We have presented a new generative model for landscape animations derived from StyleGAN, and have shown that it can be trained from the mixture of static images and timelapse videos, benefiting from both sources. We have investigated how the resulting model can be used to bring to life (reenact) static landscape images, and have shown that this can be done more successfully than with previously proposed methods. Extensive results of our method are shown in the supplementary video. 

The supplementary video also shows failure modes. Being heavily reliant on machine learning, our approach fails when reenacting static images atypical for its training dataset. Furthermore, as our video dataset is relatively small and focuses on slower motions (clouds), we have found that method often fails to animate waves and grass sufficiently strongly or realistically. Enlarging the image dataset and, in particular, the video dataset seems to be the most straightforward way to address these shortcomings.

\FloatBarrier
\ifnum\value{page}>14 \errmessage{Number of pages exceeded!!!!}\fi

\bibliographystyle{splncs}
\bibliography{refs}
\newcommand{\Sallh}{{\widehat{\Sall}}}
\newcommand{\Sdh}{{\widehat{\Sd}}}
\newcommand{\Ssh}{{\widehat{\Ss}}}

\newcommand{\Wallh}{{\widehat{\Wall}}}

\part*{Appendix}
\label{appendix}




\section{Training the main model}

\subsubsection{Training Configuration}
Our final models follows original StyleGAN training schedule. We alternate between two phases: a resolution transition phase for 600k samples then a stabilization phase for 600k samples. After final reslution is reached we train model until the number of batches reaches 450k.
The proportion of pairwise discriminator changes linearly from 0.5 to 0.1 during the resolution transition phase. 
We use crops instead of generated frames when update pairwise discriminator with probabily 0.5.
For inference we used accumulated exponential moving average with $\alpha=0.999$ to generate samples.
Our final model was trained using Adam optimizer with parameters $\beta_1=0, \beta_2=0.99$.

As in the original StyleGAN, we change batch size parameter depending on resolution: ($4\text{px}$, $512$), ($8\text{px}$, $256$), ($16\text{px}$, $128$), ($32\text{px}$, $64$), ($64\text{px}$, $32$), ($128\text{px}$, $32$), ($256\text{px}$, $16$), ($512\text{px}$, $8$), ($1024\text{px}$, $8$). Learning rates are: (up to $128\text{px}$, 1e-3), ($128\text{px}$, 1.5e-3), ($256\text{px}$, 2e-3), (eq. or bigger than $256\text{px}$, 3e-3)

\subsubsection{Pairwise Discriminator}
The pairwise discriminator differs from the original StyleGAN discriminator only in the input Conv 1x1 layer which has half the number of output channels of the original StyleGAN discriminator and is applied to each frame independently. After that both feature maps are concatenated.

\subsubsection{Balancing discriminators}
To choose the most effective way of balancing two discriminators we evaluated four different experiments (image resolution is 128px). 
While $\text{freq}=0.3$ and $\text{freq}=0.5$ suffer from much worse image quality, \textit{decay} and $\text{freq}=0.1$ behaves similarly but \textit{decay} works slightly better on moving objects and generates more compelling dynamics.

\begin{table}
    \centering
    \begin{tabular}{c|c c c c}
        \hline\hline
        Setup           & FID$\downarrow$   & SSIM$\uparrow$& LPIPS$\downarrow$ & R$\uparrow$ \\
        \hline
        \textit{decay} & 56.13 & 0.884 & 0.062 &  \textbf{0.00}  \\ 
        \hline        
        freq = 0.1 & \textbf{54.15} & 0.880 & \textbf{0.064} & -0.03\\
        freq = 0.3 & 63.48 & 0.887 & 0.058 & -0.12\\
        freq = 0.5 & 82.10 & \textbf{0.893} & 0.055 & -0.15\\
    \end{tabular}
    \caption{Different techniques to balance discriminators.
The column R in the table is obtained from the side-by-side user study. It shows the change in frequency when assessors prefer this variant to the default one (\textit{decay}).
Although decaying the relative frequency doesn't give the best results when comparing against any quantitative metric, it balances image quality with motion plausibility and wins user preference.
    }
\end{table}

\section{Inference Details}

Our overall inference procedure consists of the following steps.
\begin{enumerate}
    \item Training encoder $\E$ on a dataset of samples from a pretrained $\G$.
    \item Given a real image $x$ to be animated, obtain a set of style vectors $\Wall'$ using $\E$.
    \item Starting from $\Wall'$, find $\Wallh$ and $\Sallh$ with gradient descent, to improve reconstruction and preserve ability to animate.
    \item Having $x, \Wallh, \Sallh$ fixed, optimize $\G$ to improve reconstruction even more.
\end{enumerate}

Steps 1, 2, 4 are pretty simple, so their description in the paper is fairly detailed. Thus, here we present only the extended definition of the step 3 (latents optimization). We present two variants: one without using a segmentation mask (Algorithm~\ref{algo:eoi}, a part of EOI and EOIF); and another one relying on a segmentation mask to route information between $\Ss$ and $\Sd$ (Algorithm~\ref{algo:eoifs}, a part of EOIFS). Other variants of inference can be obtained by changing EOI, specifically:
\begin{itemize}
    \item turning off the $\Wall$ penalty $L_{init}^O$ (EO, MO, I2S);
    \item changing $\Wall$ initialization: mean style instead of $\E$ predictions (MO, I2S);
    \item changing $\Sall$ initialization: random instead of zero (I2S, E);
    \item turning off optimization of $\Sall$ (I2S);
    \item not optimizing latents at all (E).
\end{itemize} 

In I2S, we also tried using $\E$-based initialization for $\Wall$, with no success. Initialization of $\Sall$ is not very important in EOI, EOIF, EOIFS, but starting from zeros slightly helps stability.

In order to regularize $\Sall$ and to prevent too much details to be described by spatial inputs, we tried both L2-regularization and gradient scaling. While L2 helps, we found gradient scaling much more efficient: it leads to better convergence (more accurate reconstruction) and still allows to push information from $\Sall$ to $\Wall$. We found experimentally that during latents optimization $\frac{\partial L^O} {\partial \Sall}$ should be divided by 1000 for best results. This effectively changes relative learning rate for $\Sall$, comparing to the learning rate of $\Wall$.

\begin{algorithm}
\caption{EOI: initialize with \textbf{E}ncoder, \textbf{O}ptimize, tie $\Wall$ to \textbf{I}nitial values}
\label{algo:eoi}

\textbf{Inputs:} generator $\G$, style initialization $\Wall'$, input image $x$.

\textbf{Outputs:} optimized $\Wallh, \Sallh$

\textbf{Hyperparameters:} number of iterations $N$, perceptual loss coefficient $\lambda_{PL} = 0.01$, gradient scale for $\Sall$ $\lambda_{\Sall} = 0.001$, Adam learning rate $lr=0.1$.

\begin{algorithmic}[1]
    \State $\Wallh \gets \Wall'$
    \State $\Sallh \gets 0$
    \State $UpdateRule \gets$ initialize Adam optimizer for $\Wallh, \Sallh$
    \State $iter \gets 0$
    \While{$iter < N$}
      \State $y \gets \G(\Wallh, \Sallh)$ \Comment{Obtain reconstructed image}
      \State $L^O_{rec} \gets MAE(y, x) + \lambda_{PL} PL(y, x)$ \Comment{Reconstruction loss}
      \State $L^O_{init} \gets MSE(\Wallh, \Wall')$ \Comment{Style regularization}
      \State $L^O \gets L^O_{rec} + L^O_{init}$ \Comment{Total latents loss}
      \State Calculate $\frac{\partial L^O} {\partial \Wallh, \Sallh}$ \Comment{loss.backward()}
      \State $\frac{\partial L^O} {\partial \Sallh} \gets \lambda_\Sall \frac{\partial L^O} {\partial \Sallh}$ \Comment{Scale gradients for $\Sallh$}
      \State $\Wallh, \Sallh \gets UpdateRule(\Wallh, \Sallh, \frac{\partial L^O} {\partial \Wallh, \Sallh})$
      
      \State If $L^O$ does not improve over 20 iterations, halve $lr$
      \State If $L^O$ does not improve over 100 iterations, stop early
      \State $iter \gets iter + 1$
    \EndWhile
    \State \textbf{return} $\Wallh, \Sallh$
\end{algorithmic}
\end{algorithm}

\begin{algorithm}
\caption{EOIFS: initialize with \textbf{E}ncoder, \textbf{O}ptimize, tie $\Wall$ to \textbf{I}nitial values, guide $\Sall$ with \textbf{S}egmentation}
\label{algo:eoifs}

\textbf{Inputs:} generator $\G$, style initialization $\Wall'$, input image $x$, static regions mask $m$ (1 for static regions, 0 for sky and water).

\textbf{Outputs:} optimized $\Wallh, \Sallh$

\textbf{Hyperparameters:} number of iterations $N$, perceptual loss coefficient $\lambda_{PL} = 0.01$, gradient scale for $\Sall$ $\lambda_{\Sall} = 0.001$, Adam learning rate $lr=0.1$.

\begin{algorithmic}[1]
    \State $\Wallh \gets \Wall'$
    \State $\Ssh, \Sdh \gets 0$ \Comment{Initialize $\Sallh$ with zeros}
    \State $UpdateRule \gets$ initialize Adam optimizer for $\Wallh, \Sallh$
    \State $iter \gets 0$
    \While{$iter < N$}
      \State $y \gets \G(\Wallh, \Sallh)$ \Comment{Obtain reconstructed image}
      \State $L^O_{init} \gets MSE(\Wallh, \Wall')$ \Comment{Style regularization}

      \If{iter \% 2 == 0} \Comment{Even iterations are for static regions}
        \State $y_m \gets y \circ m$  \Comment{Zero out dynamic regions}
        \State $x_m \gets x \circ m$
        \State $L^O_{rec} \gets MAE(y_m, x_m) + \lambda_{PL} PL(y_m, x_m)$ \Comment{Reconstruction loss}
        \State $L^O \gets L^O_{rec} + L^O_{init}$ \Comment{Total latents loss}
        \State Calculate $\frac{\partial L^O} {\partial \Wallh, \Ssh}$ \Comment{Calculate grad only w.r.t $\Ssh$}
        \State $\frac{\partial L^O} {\partial \Ssh} \gets \lambda_\Sall \frac{\partial L^O} {\partial \Ssh}$ \Comment{Scale gradients for $\Ssh$}
        \State $\Wallh, \Ssh \gets UpdateRule(\Wallh, \Ssh, \frac{\partial L^O} {\partial \Wallh, \Ssh})$
      \Else
        \State $y_m \gets y \circ (1 - m)$  \Comment{Zero out static regions}
        \State $x_m \gets x \circ (1 - m)$
        \State $L^O_{rec} \gets MAE(y_m, x_m) + \lambda_{PL} PL(y_m, x_m)$ \Comment{Reconstruction loss}
        \State $L^O \gets L^O_{rec} + L^O_{init}$ \Comment{Total latents loss}
        \State Calculate $\frac{\partial L^O} {\partial \Wallh, \Sdh}$ \Comment{Calculate grad only w.r.t $\Sdh$}
        \State $\frac{\partial L^O} {\partial \Sdh} \gets \lambda_\Sall \frac{\partial L^O} {\partial \Sdh}$ \Comment{Scale gradients for $\Sdh$}
        \State $\Wallh, \Sdh \gets UpdateRule(\Wallh, \Sdh, \frac{\partial L^O} {\partial \Wallh, \Sdh})$
      \EndIf
      \State If $L^O$ does not improve over 20 iterations, halve $lr$
      \State $iter \gets iter + 1$
    \EndWhile
    \State \textbf{return} $\Wallh, \Sallh$
\end{algorithmic}
\end{algorithm}

\subsection{What $\zd$ Actually Describe?}

In order to manipulate lighting on a real image, we train a dedicated neural network $\A$, which approximates local dynamics of a multilayer perceptron $\M$, which maps $\zall$ to $\w$. During training of $\G$ and $\M$, $\zd \in \mathbb{R}^3$ is sampled from standard normal distribution. However, it is not practical to sample styles for real images, because we usually want to get something concrete (e.g. day to evening or evening to night conversion).

Thus, we needed a technique to build an "interpretation" of 3 numbers which make up $\zd$. A well established approach for that is to (a) sample a set of synthetic images from $\G$, (b) manually assign them class labels (e.g. day, evening, night); (c) obtain "direction vectors", which correspond to the shortest path from one class to another in the latent space. Having direction vectors, one can modify $\zd$ along them in order to change image style accordingly. This approach can help to build an interpretation of a complex high-dimensional model.

However, in our case we have only 3 components to interpret, thus we decided to take a more simple way: manually change $\zd$ coordinates one-by-one and try to describe the way the image changes. For each coordinate we tried values from $\{-3, -2, -1, 0, 1, 2, 3\}$ while keeping other coordinates zero. We also tried changing pairs and triplets of coordinates the same way.

We found that as a result of multiple $\G$ training sessions on the same dataset, $\zd$ consistently received approximately the following semantics:
\begin{enumerate}
    \item The first coordinate changes brightness without altering color temperature (day-to-night). Thus, when moving from day to night we do not arrive to a warm yellow sunset.
    \item The second coordinate changes brightness and color temperature together: negative values lead to darker images with all lights (city, sunset) getting more saturated and warm; positive values lead to lighter images with colder colors. One can get day-to-evening conversion with that coordinate alone.
    \item The third coordinate does almost the same as the first one does. We found no significant difference between them.
    \item By changing the first and the second coordinates, one can obtain dark night, warm sunset, blue hour, bright day with clouds, bright day with clear sky.
\end{enumerate}

Our experiments show that $\zd$ affects image style in a fairly monotonic way.

Using the described methodology, we constructed a vocabulary of 9 styles, which correspond to different time of day and weather. We use only these styles for all videos where we animate real images for our quantitative and qualitative experiments. We use styles randomly sample from normal distribution for fully synthetic videos.

\section{Inference procedure ablation study.}
\label{infquant}
Where we quantify the impact of different elements of our inference algorithm on the reconstruction accuracy, image quality, static consistency and motion amount. Image quality and static consistency for best inference variants (EOIF and EOIFS) are discussed in the paper, Section 4 (Experiments). The reconstruction quality is evaluated via LPIPS and SSIM measured between the input and the reconstructed images. The amount of motion is quantified as mean optical flow~\cite{jiang2018super} in the sky region, according to semantic segmentation. We generate videos in the same way as for other experiments. The results of this ablation study (\tab{inference-ablation-metrics}) verify that all steps of our inference procedure are needed to obtain animations that both have plausible motion and fit the input images well.

\begin{table}
    \centering

    \begin{tabular}{c|c|c|c}
        \hline\hline
        Algorithm                            & SSIM$\uparrow$ & LPIPS$\downarrow$ & Flow$ \times 10^{-3} \uparrow$ \\
        \hline\hline
        I2S~\cite{Abdal2019Image2StyleGANHT} & 0.80 & 0.18 & 1.3 \\
        MO                                   & 0.95 & 0.07 & 1.3 \\
        E                                    & 0.54 & 0.43  & 2.2 \\
        EO                                   & 0.95 & 0.07 & 1.5 \\
        EOI                                  & 0.92 & 0.11 & 2.5 \\
        EOIF                                 & \textbf{0.96} & \textbf{0.04} & 2.3 \\
        EOIFS                                & 0.94 & 0.05 & \textbf{2.6} \\
        \hline\hline
    \end{tabular}
    
    \caption{Quantitative evaluation of inference procedure: reconstruction quality and motion amount. While sevral variants result in good reconstruction, only EOIF and EOIFS variants yield both good reconstruction and motion.}
    \label{tab:inference-ablation-metrics}
\end{table}

\section{Animated Landscape finetuning}
In order to ensure fair comparison, we tried to reproduce results from AL paper using only our video dataset. We tried both training from scratch and finetuning the publicly available model for 50 to 200 epochs. Training from scratch did not converge, so we present here only metrics obtained with finetuning. Table~\ref{tab:finetuned-al} shows that finetuning damages the model. This is most probably due to the very small dataset, which contains motions of very different speed. Authors of AL somehow equalized speed of different videos, but the exact methodology for that is unknown. Training on unequalized videos is harmful. On contrary, our model does not require equalization of motion speed, which allows to use more dirty data without degradation of performance.

\begin{table}
    \centering

    \begin{tabular}{c|c| c |c |c| c}
        \hline\hline
        & FVD$\downarrow$  & LPIPS$\downarrow$ & SSIM$\uparrow$ & FID$\downarrow$ & User preference$\uparrow$ \\
        \hline\hline
        AL$_\text{noint}$ & 162 & \textbf{0.063} & 0.92 & \textbf{51.9}  & \textbf{0.68} \\
        AL$_\text{finetuned}$ & \textbf{159} & 0.065 & 0.92 &  53.4 & 0.32 \\
        \hline\hline
    \end{tabular}
    
    \caption{Comparison of pretrained and finetuned AL}
    \label{tab:finetuned-al}
\end{table}


\begin{figure}
    \centering
    \includegraphics[width=0.5\textwidth]{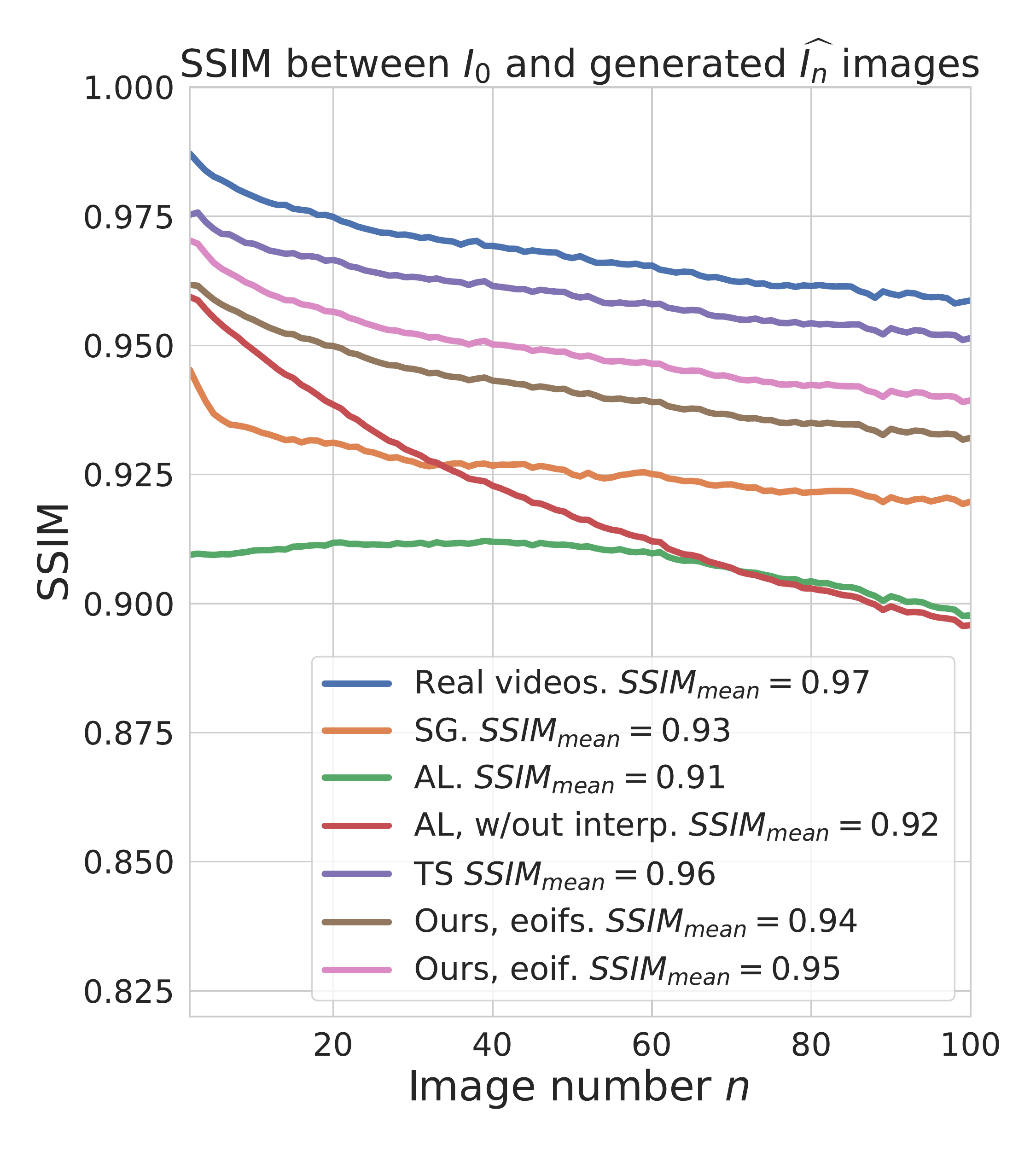}
    \caption{Continuation of Figure 7. Quantitative comparison of image quality, static consistency and motion plausibility}
    \label{fig:ssim}
\end{figure}

\section{The Structural Similarity Index (SSIM)}
Figure~\ref{fig:ssim} presents masked SSIM between $I_0$ and $\widehat{I}_n$, which mostly measure image quality and static consistency. Note that TS baseline, which uses segmentation and simply copies static parts, outperforms other methods (but losses the game when it comes to perceptual quality and motion plausibility).

\section{Side-by-side Comparison on the speed of real videos}

\begin{figure}
    \centering
         \begin{tabular}{l|c|c}
            \hline\hline
            \textbf{Method} 
            & \textbf{EOIF} & \textbf{EOIFS} \\
            \hline\hline
            SG & 0.28 & 0.27
            \\ \hline
            AL (no int) & 0.33 & 0.36
            \\ \hline
            AL (+ style) & 0.14 & 0.12
            \\ \hline
            TS & 0.11 & 0.12
            \\ \hline
            Real & 0.68 & 0.70
            \\ \hline
            Ours (EOIF) & -- & 0.52 
            \\ \hline
            Ours (EOIFS) & 0.48 & -- \\
            \hline\hline
        \end{tabular}   
    \caption{Ratio of wins row-over-column for side-by-side setting B, synthetic video speed aligned to that of real ones (faster videos).}
    \label{fig:real-speed-user-study}
\end{figure}

On Figure~\ref{fig:real-speed-user-study} we present side-by-side user study, setup B, i.e. with speed of synthetic videos aligned to that of real ones. Note that real videos win more often, but advantage of our method against competitors is even more evident.
\end{document}